\documentclass[11pt,a4paper]{article}
\usepackage{fix-cm}
\usepackage{amsmath,amssymb}
\usepackage{graphicx}
\usepackage{booktabs,array,tabularx,longtable}
\usepackage[table]{xcolor}
\usepackage{caption}
\usepackage{enumitem}
\usepackage[numbers,sort&compress]{natbib}
\usepackage{url}
\usepackage{report}
\usepackage[colorlinks=true,linkcolor=reportblue,citecolor=reportblue,urlcolor=reportblue]{hyperref}
\hypersetup{pdftitle={LaBraM at Scale: Scaling Data and Models for EEG Pretraining},pdfauthor={Jun-Yu Pan, Ruicheng Yin, Bao-Liang Lu, Dongsheng Li, Wei-Long Zheng, Yansen Wang},pdfsubject={EEG pretraining, model architecture, training infrastructure, and transfer evaluation},pdfstartpage=1,pdfstartview=Fit,pdfpagemode=UseNone,pdfpagelayout=OneColumn}
\graphicspath{{figures/}}
\newcommand{\reportauthors}{%
  {\fontsize{11}{14}\selectfont
    \textbf{Jun-Yu Pan}\textsuperscript{1,}\footnotemark[2]\qquad
    \textbf{Ruicheng Yin}\textsuperscript{2}\qquad
    \textbf{Bao-Liang Lu}\textsuperscript{1}\qquad
    \textbf{Dongsheng Li}\textsuperscript{3}\par
    \vspace{2pt}
    \textbf{Wei-Long Zheng}\textsuperscript{1,}\footnotemark[1]\qquad
    \textbf{Yansen Wang}\textsuperscript{3,}\footnotemark[1]\par}
  \vspace{4pt}
  {\fontsize{9.5}{11.5}\selectfont
    \textsuperscript{1}Shanghai Jiao Tong University, Shanghai, China\quad
    \textsuperscript{2}Fudan University, Shanghai, China\par
    \textsuperscript{3}Microsoft Research Asia, Shanghai, China\par}
  \vspace{3pt}
  {\fontsize{8.5}{10.5}\selectfont
    \texttt{\{panjunyu,bllu,weilong\}@sjtu.edu.cn}\quad
    \texttt{rcyin23@m.fudan.edu.cn}\quad
    \texttt{\{dongsli,yansenwang\}@microsoft.com}\par}
}
\newcommand{\tabnote}[1]{\par\vspace{3pt}{\footnotesize\raggedright #1\par}}
\newcommand{\na}{\textemdash}
\newenvironment{reportfigure}{%
  \begin{figure}[!htbp]\centering\captionsetup{hypcap=false}%
}{\end{figure}}
\newenvironment{reporttable}{%
  \par\addvspace{8pt}\noindent\begin{minipage}{\linewidth}%
  \captionsetup{type=table,hypcap=false}\centering
}{\end{minipage}\par\addvspace{8pt}}

\begin{document}
\thispagestyle{opening}
\vspace*{0pt}
\begingroup\centering\setlength{\parskip}{0pt}
{\resizebox{\linewidth}{!}{\bfseries LaBraM at Scale: Scaling Data and Models for EEG Pretraining}\par}
\vspace{8pt}
\begingroup\renewcommand{\thefootnote}{\fnsymbol{footnote}}
\reportauthors
\endgroup
\par\endgroup
\begingroup\renewcommand{\thefootnote}{\fnsymbol{footnote}}
\footnotetext[1]{Corresponding authors.}
\footnotetext[2]{Work done during Jun-Yu Pan's internship at Microsoft Research Asia.}
\endgroup
\vspace{6pt}
{\centering\textbf{Abstract}\par}
\vspace{3pt}
\begingroup\small\noindent
LaBraM, an early EEG foundation model, demonstrated that large-scale EEG pretraining can yield strong representations for diverse downstream tasks. Building on this foundation, we expand pretraining data and model capacity while retaining its core encoder and objective. We train ten encoders with 5.8 million to 1.97 billion parameters on three EEG collections spanning 25,067--79,910 hours, adapting signal preparation and training to heterogeneous recordings. Across 23 tasks, \mbox{Atom-L$^*$} reaches 59.05\% mean balanced accuracy at its best evaluated full fine-tuning (Full-FT) checkpoint, 3.61 percentage points above the strongest baseline EEG foundation model. Quark-G leads linear probing (LP) at 54.33\%, a 1.24-point margin. Both also lead average task rank. At the last evaluated checkpoints, larger encoders improve Full-FT more consistently than LP, and stronger LP does not guarantee stronger Full-FT. Preserving patch structure improves LP over average pooling. Data-fraction experiments suggest capacity-dependent saturation, while checkpoint trajectories show that downstream performance can decline as pretraining loss falls. Joint classification with a shared Qwen3 readout reaches 50.50\% with Molecule-H and Qwen3-8B; increasing language-model size provides no consistent gain. These results demonstrate the value of scaling LaBraM and the importance of matching pretraining scale to downstream adaptation.
\par\endgroup
\vspace{6pt}
\noindent\begin{minipage}{\linewidth}
\centering
\includegraphics[width=0.74\linewidth]{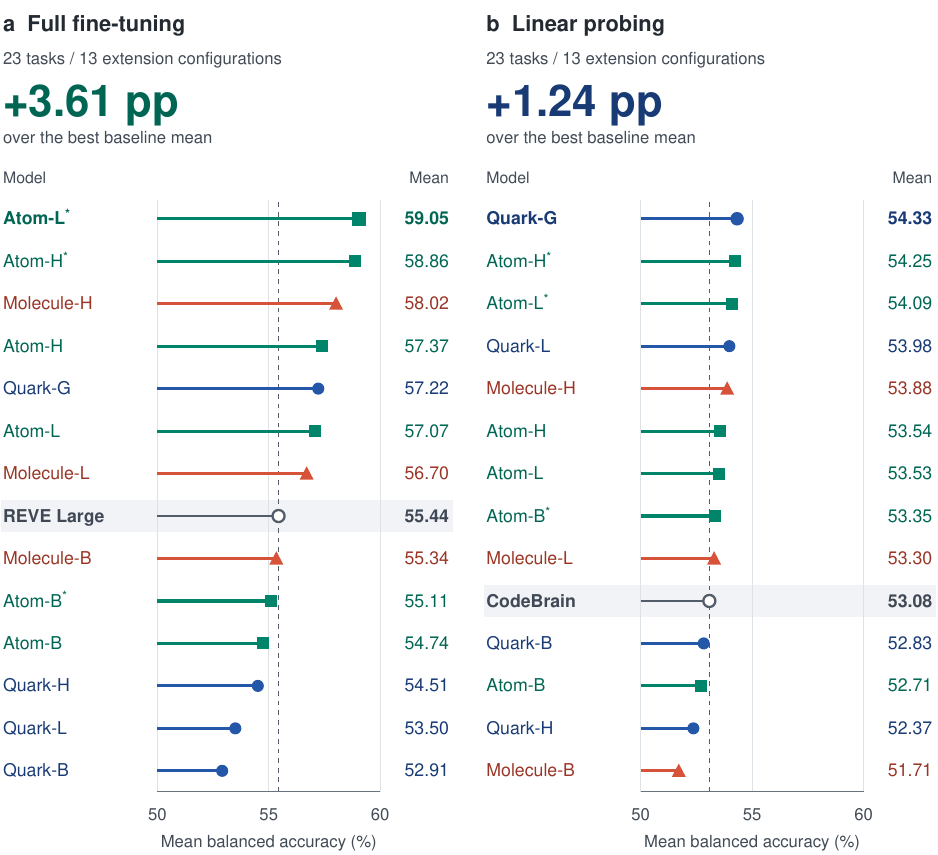}
\captionsetup{hypcap=false}
\captionof{figure}{Full-FT and LP performance of ten extension encoders and three selected-checkpoint references ($^*$) on 23 tasks. Stars mark checkpoints selected by mean Full-FT score; LP uses the same checkpoints. Dashed lines mark the strongest baseline in each setting. Figures~\ref{fig:model-comparison} and~\ref{fig:lp-rankings} show the complete comparisons.}
\label{fig:headline}
\end{minipage}
\section{Introduction}
\label{sec:overview}

LaBraM, an early EEG foundation model, demonstrated that pretraining on large EEG collections can learn representations that transfer effectively across diverse downstream tasks~\cite{jiang2024labram}. Subsequent EEG foundation models have provided further evidence for the effectiveness of large-scale pretraining~\cite{wang2025cbramod,doner2025luna,elouahidi2025reve,jiang2025neurolm,tegon2025femba,wang2025eegmamba,xiao2025brainomni,zhou2025csbrain,ma2026codebrain}. Building on LaBraM, we further investigate the downstream gains from expanding pretraining data and model capacity while retaining its core encoder design and pretraining objective.

To study this expansion, we assemble three EEG collections---Quark, Atom, and Molecule---spanning 25,067--79,910 hours, and train ten encoders ranging from 5.8 million to 1.97 billion parameters. Combining recordings from diverse sources requires handling differences in channel layout, signal scale, and recording length. Extended signal preparation and batching accommodate these inputs, while streamed data access, node-local storage, and distributed training support the larger collections and encoders.

The resulting encoders lead the comparisons across 23 classification tasks under both full fine-tuning (Full-FT) and linear probing (LP). With 46.3 million parameters, \mbox{Atom-L$^*$} reaches 59.05\% mean balanced accuracy at its best evaluated Full-FT checkpoint, exceeding REVE Large by 3.61 percentage points. Quark-G leads LP at 54.33\%, and both models lead average task rank in their respective settings. The gains also extend to comparisons with small supervised models: Atom-L and Molecule-H each exceed the stronger of EEGNet and EEGConformer on 16 tasks under Full-FT, compared with at most 12 for a baseline EEG foundation model (Section~\ref{sec:evaluation}).

Further analysis examines how downstream gains vary with encoder size, pretraining data, and training duration. At the last evaluated checkpoints, larger encoders improve Full-FT more consistently than LP. The two evaluation settings also produce different model orderings: stronger LP does not guarantee stronger Full-FT. Retaining patch structure improves LP over average pooling, showing that the readout affects how well frozen features can be used. Together, these results motivate assessing both access to frozen features and the gains from task-specific encoder adaptation (Section~\ref{sec:linear-probing}).

The data experiments show why recording hours alone do not determine transfer performance. Reduced-data Atom runs outperform Quark at matched encoder sizes despite using fewer hours, suggesting that broader source coverage may improve data efficiency. Within a fixed collection, additional data can eventually bring diminishing or negative returns. Fraction sweeps motivate an empirical capacity-dependent saturation relation, with fitted Full-FT peaks near 22,000 hours for Base and 40,000 hours for Large. Even within a single training run, downstream performance can decline while pretraining loss falls. Judging the benefit of further pretraining therefore requires downstream evaluation (Section~\ref{sec:observations}).

Beyond traditional evaluation, we further explore our models' performance in a multitask setting, drawing inspiration from NeuroLM~\cite{jiang2025neurolm}, UniMind~\cite{lu2025unimind}, and GVG~\cite{pan2026gvg}. Frozen EEG encoders are paired with a shared Qwen3 readout for joint classification across all 23 tasks. Larger EEG collections and encoders show complementary gains, with Molecule-H and Qwen3-8B reaching 50.50\% mean balanced accuracy. Increasing language-model size provides no consistent improvement. Together with the LP analysis, these results suggest that preserving EEG information and adapting it to the readout are important for realizing the benefits of encoder scale (Section~\ref{sec:eeg-llm}).

\section{Pretraining Data}
\label{sec:data}

\subsection{From Quark to Atom and Molecule}

Our pretraining experiments use three data versions (Table~\ref{tab:data-scales}): Quark comprises 9 datasets, 25,067 hours, and 4,200,222 training windows; Atom comprises 521 datasets, 49,628 hours, and 24,723,392 windows; and Molecule comprises 505 datasets, 79,910 hours, and 35,207,959 windows.

\begin{reporttable}
\centering
\small
\caption{Pretraining data summary.}
\label{tab:data-scales}
\begin{tabular}{@{}lrrr@{}}
\toprule
Data version & Datasets & Hours & Training windows \\
\midrule
Quark & 9 & 25,067 & 4,200,222 \\
Atom & 521 & 49,628 & 24,723,392 \\
Molecule & 505 & 79,910 & 35,207,959 \\
\bottomrule
\end{tabular}
\end{reporttable}

Quark forms the initial collection and is dominated by clinical recordings: TUH contributes 23,653 hours. The remaining sources are HMC, Alljoined, THINGS-EEG2, a self-collected dataset, SEED, SEED-IV, SEED-V, and SEED-VII.

To explore the limits of data scaling, we assembled Atom and Molecule by collecting publicly available EEG datasets as comprehensively as possible and screening them for pretraining. Atom broadens this clinical starting point to include more recording settings and tasks. Added sources include I-CARE, CHB-MIT, Siena, ISRUC S3, FACED, EEGMMIDB, SparrKULee, and THINGS-EEG1, together with additional collections from OpenNeuro, Zenodo, and MOABB.

Molecule builds on this broader source pool with revised signal preparation and source selection. For this expansion, we collected approximately 92,032 hours of EEG from 514 datasets and constructed 41,637,575 windows before quality screening. Table~\ref{tab:pretraining-named-sources} compares recording hours for selected sources across all three versions; Appendix~\ref{app:pretraining-datasets} provides the complete inventory with publication and dataset links.

\begingroup
\small
\setlength{\LTpre}{6pt}
\setlength{\LTpost}{8pt}
\begin{longtable}{@{}l>{\raggedright\arraybackslash}p{0.24\linewidth}rrr@{}}
\caption{Selected pretraining datasets, sources, and recording hours. The full inventory appears in Table~\ref{tab:pretraining-inventory}.}\label{tab:pretraining-named-sources}\\
\toprule
Dataset & Source & Quark (h) & Atom (h) & Molecule (h) \\
\midrule
\endfirsthead
\caption[]{Selected pretraining datasets, sources, and recording hours. The full inventory appears in Table~\ref{tab:pretraining-inventory} (continued).}\\
\toprule
Dataset & Source & Quark (h) & Atom (h) & Molecule (h) \\
\midrule
\endhead
\midrule
\multicolumn{5}{r@{}}{\emph{Continued on the next page.}}\\
\endfoot
\bottomrule
\endlastfoot
TUH & \cite{dataset1ee20e82d63f} & 23,653 & 22,554 & 22,158 \\
I-\allowbreak{}CARE & \cite{dataset1a320341e502} & -- & 1,657 & 28,583 \\
CHB-\allowbreak{}MIT & \cite{dataset9411f28d9da7} & -- & 991 & 1,964 \\
HMC & \cite{dataset8c058cac92db} & 741 & 17 & 1,142 \\
ON\_\allowbreak{}ds004395 & \cite{dataset6ab9992ffa53} & -- & 729 & 6,827 \\
ON\_\allowbreak{}ds005555 & \cite{dataset59cc2be65a23} & -- & 2,002 & 1,947 \\
Alljoined & \cite{datasetf59c61126677} & 371 & -- & 461 \\
THINGS-\allowbreak{}EEG2 & \cite{dataset822d9c0bbdd7} & 184 & -- & 228 \\
self-\allowbreak{}collected dataset &  & 28 & 28 & 22 \\
SEED & \cite{dataset0678103fbd8b} & 16 & 9 & -- \\
SEED-\allowbreak{}IV & \cite{dataset74f7ed3a6339} & 18 & 9 & 18 \\
SEED-\allowbreak{}V & \cite{dataset099fab2a0498} & 13 & 9 & 13 \\
SEED-\allowbreak{}VII & \cite{datasetd169eca259ba} & 43 & 9 & 43 \\
\end{longtable}
\tabnote{A dash denotes absence from a collection.}
\endgroup

This source-level comparison shows that most of Molecule's additional recording time comes from ICU data. I-CARE increases from 1,657 hours in Atom to 28,583 hours in Molecule, accounting for about 89\% of the net increase between the two versions. Molecule therefore combines a larger ICU component with cognitive, sleep, and other experimental recordings from the broader source pool (Figure~\ref{fig:data-composition}). Appendix~\ref{app:additional-settings} gives the category totals behind this change in composition.

\begin{reportfigure}
\includegraphics[width=\linewidth]{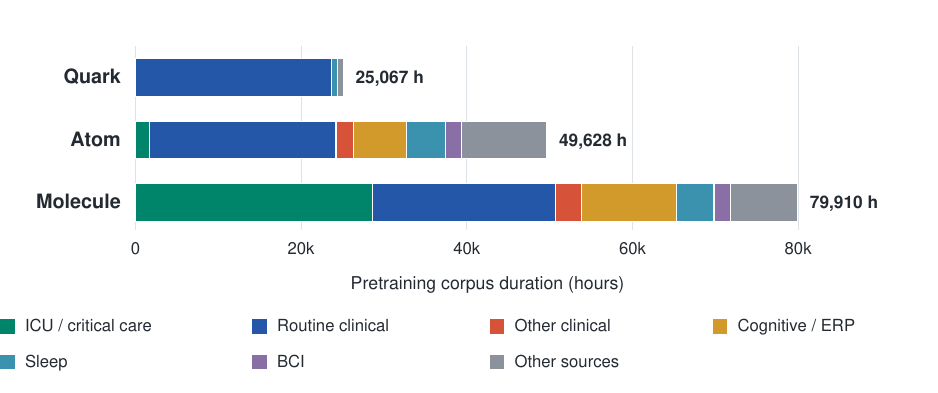}
\caption{Pretraining data composition. Categories group sources by their predominant recording setting or task and are approximate. Molecule's increase in duration is concentrated in ICU recordings.}
\label{fig:data-composition}
\end{reportfigure}

\subsection{Signal preparation and window construction}

Public EEG datasets use different sampling rates and electrode naming conventions. We resample recordings to 200 Hz, apply band-pass filtering with a 0.5 Hz lower cutoff, and map electrode names to a shared vocabulary for the encoder.

Recordings can also contain unusable channels and gaps. Channels that are constant or contain at least 1\% non-finite samples are removed; the remaining signals are split into finite segments to retain usable EEG around gaps. Per-channel z-scoring within each recording or trial reduces amplitude scale differences across sources. Values are then clipped to $\pm15$ and stored in float16. Because normalization is performed during preparation, we use an input scale of 1.0; Quark and Atom retain 0.01 (Appendix~\ref{app:molecule-preparation}).

With a fixed window duration, recordings with more channels produce more tokens and require more attention memory. We therefore use non-overlapping windows of 4--16 seconds, adjusting duration to target 256 channel-by-time tokens. A 16-channel recording uses 16-second windows, while 64 channels use four seconds. Alljoined and THINGS-EEG2 retain their one-second trials to preserve trial boundaries. Tails of at least one second are zero-padded and retained with validity masks, preserving available EEG while marking padded positions.

\subsection{Quality control and sampling}

After preprocessing, windows still differ in signal quality and channel coverage. We remove windows with severe signal abnormalities or excessive padding to limit training on unreliable inputs, and apply the subject exclusion list separately. For Molecule, subject exclusions remove 687,187 windows and severe-quality screening removes 5,742,429, leaving 35,207,959 of the 41,637,575 constructed windows eligible for pretraining (Figure~\ref{fig:data-preparation}).

We retain windows with mild quality flags or few channels because they still contain usable EEG, using lower sampling weights to favor the main pool. The retained pool comprises 26,102,536 main windows with at least eight mapped channels, 8,180,196 mildly flagged windows, and 925,227 windows with fewer than eight mapped channels.

\begin{reportfigure}
\includegraphics[width=\linewidth]{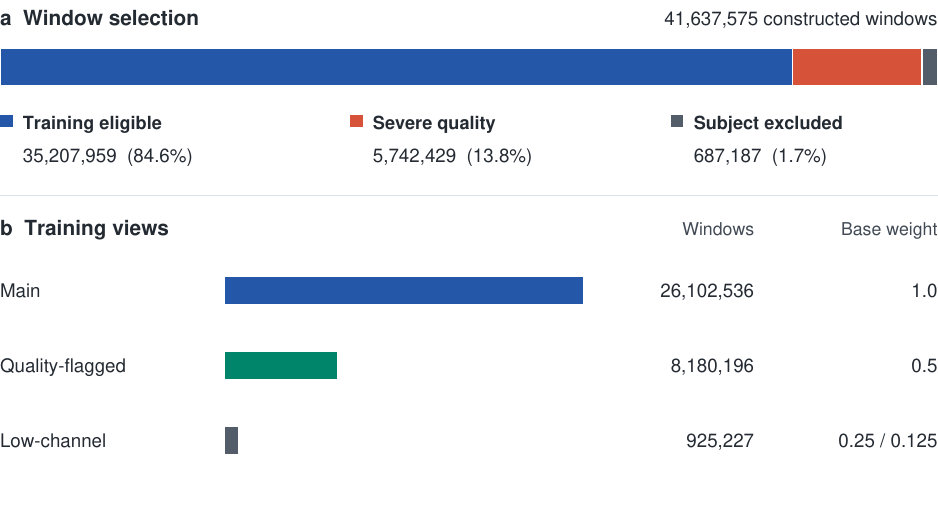}
\caption{Molecule window screening and sampling. Screening excludes 15.4\% of constructed windows; counts are shown before weighted sampling. Low-channel windows use weight 0.25, or 0.125 when they also carry quality flags. Source and provenance factors apply separately. Subject exclusions follow the earlier evaluation pool; see Appendix~\ref{app:overlap} for remaining overlap.}
\label{fig:data-preparation}
\end{reportfigure}

\section{Scaling the Training Pipeline}
\label{sec:model}

\subsection{Encoder sizes}

We retain LaBraM's neural spectrum tokenizer architecture and masked-code prediction objective \cite{jiang2024labram} while scaling the encoder from 5.8 million to 1.97 billion parameters (Table~\ref{tab:model-sizes}). Base, Large, and Huge retain the original LaBraM definitions. Giant extends the family to width 1,600 and 64 transformer layers. Model names combine the data version with B, L, H, or G for the corresponding encoder size. The training-pipeline extensions below support heterogeneous inputs and execution at larger data and model scales.

\begin{reporttable}
\centering
\small
\caption{Encoder configurations.}
\label{tab:model-sizes}
\begin{tabular}{@{}lrrrr@{}}
\toprule
Size & Width & Depth & Heads & Parameters \\
\midrule
Base & 200 & 12 & 10 & 5.820M \\
Large & 400 & 24 & 16 & 46.259M \\
Huge & 800 & 48 & 16 & 369.236M \\
Giant & 1,600 & 64 & 20 & 1.968B \\
\bottomrule
\end{tabular}
\end{reporttable}

\subsection{Batching across datasets}

Recordings from different datasets use different electrode sets and channel orders. LaBraM assigns spatial embeddings by electrode identity, so two windows with the same tensor shape can require different embedding indices. We therefore group windows in the Atom and Molecule loaders by channel count, patch count, channel signature, and ordered channel list. Samples in each batch then share the same dimensions and electrode ordering. For Molecule, we also reject unmapped pretraining channel names before batching.

Recording tails can end partway through a patch, so the resulting windows contain different amounts of observed EEG. In the Molecule encoder, sample-validity masks carry this information from window construction into attention and prediction. A fully padded patch contains no recorded EEG: we exclude it from attention keys and prediction masks and zero its output after the final normalization. Partially padded patches retain recorded samples, so we keep them in the sequence and weight their prediction loss in proportion to the valid sample counts. Appendix~\ref{app:additional-settings} describes the tokenizer's treatment of padded reconstruction targets.

\label{sec:infrastructure}

\subsection{Streaming and node-local storage}

With tens of millions of windows, opening individual sample files would incur repeated file-access overhead. We therefore package Atom and Molecule in WebDataset tar shards so workers can read many windows from each archive as a stream. The two collections use 2,885 and 291,489 shards, respectively.

Because training revisits shards across epochs, each node caches downloaded shards, allowing workers to reuse local copies on subsequent reads. A least-recently-used policy evicts shards when space is needed. SQLite tracks the cache contents, and file locks coordinate workers requesting the same shard.

Persistent workers reduce process-startup overhead between epochs. Prefetching prepares upcoming batches while the current batch is processed, and pinned host memory supports transfer of the prepared batches to the accelerator.

\subsection{Distributed training and precision}

The training stack supports distributed data parallelism and fully sharded data parallelism (FSDP), with optional activation checkpointing to reduce activation memory. Molecule-H uses full parameter sharding; Quark-G was trained on MI300X GPUs. Table~\ref{tab:pretrain-runs} lists GPU counts and global batches.

We replace the explicit attention path with PyTorch scaled dot-product attention, which selects kernels supported by the device, precision, and tensor layout. The training loop supports bf16 without loss scaling. Quark uses fp16 with GradScaler; Molecule extracts frozen-tokenizer targets in bf16.

\section{Pretraining Configuration}
\label{sec:pretraining}
\begingroup
\setlength{\intextsep}{6pt plus 1pt minus 1pt}

The three collections differ in source composition and input preparation. We therefore use independently trained tokenizer weights for each collection and freeze them during encoder pretraining, so that target codes remain fixed as the encoder is optimized. Table~\ref{tab:tokenizers} in the appendix lists the tokenizer settings and checkpoints used for target generation.

With these targets fixed, encoder optimization follows a common recipe: AdamW with weight decay 0.05, cosine learning-rate decay to $10^{-5}$, drop path 0.1, and 50\% masking. Within this recipe, Atom uses ten warmup epochs and gradient clipping at 1.0; Molecule uses five warmup epochs without gradient clipping. Table~\ref{tab:pretrain-runs} gives the GPU allocation, global batch size, peak learning rate, and epoch schedule for each encoder.

\begingroup
\small
\setlength{\LTpre}{6pt}
\setlength{\LTpost}{8pt}
\begin{longtable}{@{}lrrrr@{}}
\caption{Encoder training configurations.}\label{tab:pretrain-runs}\\
\toprule
Model & GPUs & Global batch & Peak LR & Epochs \\
\midrule
\endfirsthead
\caption[]{Encoder training configurations (continued).}\\
\toprule
Model & GPUs & Global batch & Peak LR & Epochs \\
\midrule
\endhead
\midrule
\multicolumn{5}{r@{}}{\emph{Continued on the next page.}}\\
\endfoot
\bottomrule
\endlastfoot
Quark-B & 32 & 2,048 & $5\times10^{-4}$ & 50 \\
Quark-L & 32 & 512 & $3\times10^{-4}$ & 50 \\
Quark-H & 24 & 96 & $1.5\times10^{-4}$ & 50 \\
Quark-G & 64 & 1,024 & $1\times10^{-4}$ & 30 \\
Atom-B & 8 & 1,024 & $1.5\times10^{-4}$ & 50 \\
Atom-L & 16 & 1,024 & $1.5\times10^{-4}$ & 50 \\
Atom-H & 64 & 1,024 & $1.5\times10^{-4}$ & 50 \\
Molecule-B & 32 & 2,048 & $5\times10^{-4}$ & 50 \\
Molecule-L & 32 & 1,024 & $3\times10^{-4}$ & 30 \\
Molecule-H & 32 & 512 & $1\times10^{-4}$ & 50 \\
\end{longtable}
\endgroup

The data-fraction experiments use this setup to examine how downstream transfer changes with data volume at a fixed encoder size. These runs retain 10\%, 20\%, or 50\% of the corresponding window streams under a 50-epoch schedule. Larger fractions therefore increase both the number of training windows and the number of optimization updates. Appendices~\ref{app:additional-settings} and~\ref{app:evaluation-settings} give the subsampling and evaluation settings.
\endgroup

\section{Evaluation}
\label{sec:evaluation}

\subsection{Evaluation setup}
\label{sec:eval-data}
\label{sec:selection}

Following the evaluation framework of OmniEEG-Bench~\cite{lu2026omnieegbench}, we evaluate transfer on a subset of 23 classification tasks covering clinical EEG, motor imagery, evoked responses, emotion, sleep staging, and subject traits. Full fine-tuning (Full-FT) measures performance after updating the encoder and task head; linear probing (LP) evaluates a readout from frozen features. All downstream evaluations use at most 40 training windows per subject--class pair, retaining all validation and test windows. Scores are averaged over seeds 42, 10, and 5. We compare overall performance by mean balanced accuracy, then use within-task ranks and individual task results to examine the breadth of the advantage. Appendix~\ref{app:tasks} lists task inputs and splits; Appendix~\ref{app:evaluation-settings} gives optimization settings.

The comparison includes 16 pretrained baseline encoders. The released LaBraM Base~\cite{jiang2024labram} provides a reference for scaling the original approach. The other baselines are BENDR~\cite{kostas2021bendr}, BIOT~\cite{yang2023biot}, CBraMod~\cite{wang2025cbramod}, BrainOmni~\cite{xiao2025brainomni}, NeuroGPT~\cite{cui2023neurogpt}, NeuroLM~\cite{jiang2025neurolm}, EEGMamba~\cite{wang2025eegmamba}, FEMBA Base~\cite{tegon2025femba}, CSBrain~\cite{zhou2025csbrain}, CodeBrain~\cite{ma2026codebrain}, REVE Base and Large~\cite{elouahidi2025reve}, and LUNA Base, Large, and Huge~\cite{doner2025luna}. Full-FT also includes EEGNet~\cite{lawhern2018eegnet}, EEGConformer~\cite{song2023eegconformer}, and a randomly initialized LaBraM Base as references for learning from scratch.

Unstarred extension comparisons use the last evaluated full-data checkpoints. Within each downstream fit, validation balanced accuracy selects the epoch. Seed42 test performance selects hyperparameters, which are then reused for seeds 10 and 5. For each model and task, LP reports the available Pooled or No-pool readout with the higher three-seed test mean, preferring Pooled on ties; SleepEDF uses Pooled.

\subsection{Full fine-tuning}

The scaled LaBraM encoders lead both mean balanced accuracy and average task rank under Full-FT (Figure~\ref{fig:model-comparison}). \mbox{Atom-L$^*$} reaches 59.05\%, exceeding REVE Large, the strongest baseline EEG foundation model, by 3.61 percentage points. The advantage also holds at the last evaluated checkpoints: Molecule-H reaches 58.02\%, while Atom-L achieves 57.07\% with 46.3 million parameters. Strong Full-FT performance therefore extends to the Large encoder, alongside the larger Huge models.

The task rankings show how this improvement extends across the benchmark. Extension configurations occupy the first eight positions by average task rank. \mbox{Atom-L$^*$} leads at 8.74, and Molecule-H ranks second at 9.30; the strongest baseline by this measure, REVE Base, has an average rank of 14.72. Agreement between the two rankings supports a broad transfer advantage, with the remaining task-level differences examined in Section~\ref{sec:task-differences}.

\begin{reportfigure}
\centering
\includegraphics[width=\linewidth,trim=0 43 0 6,clip]{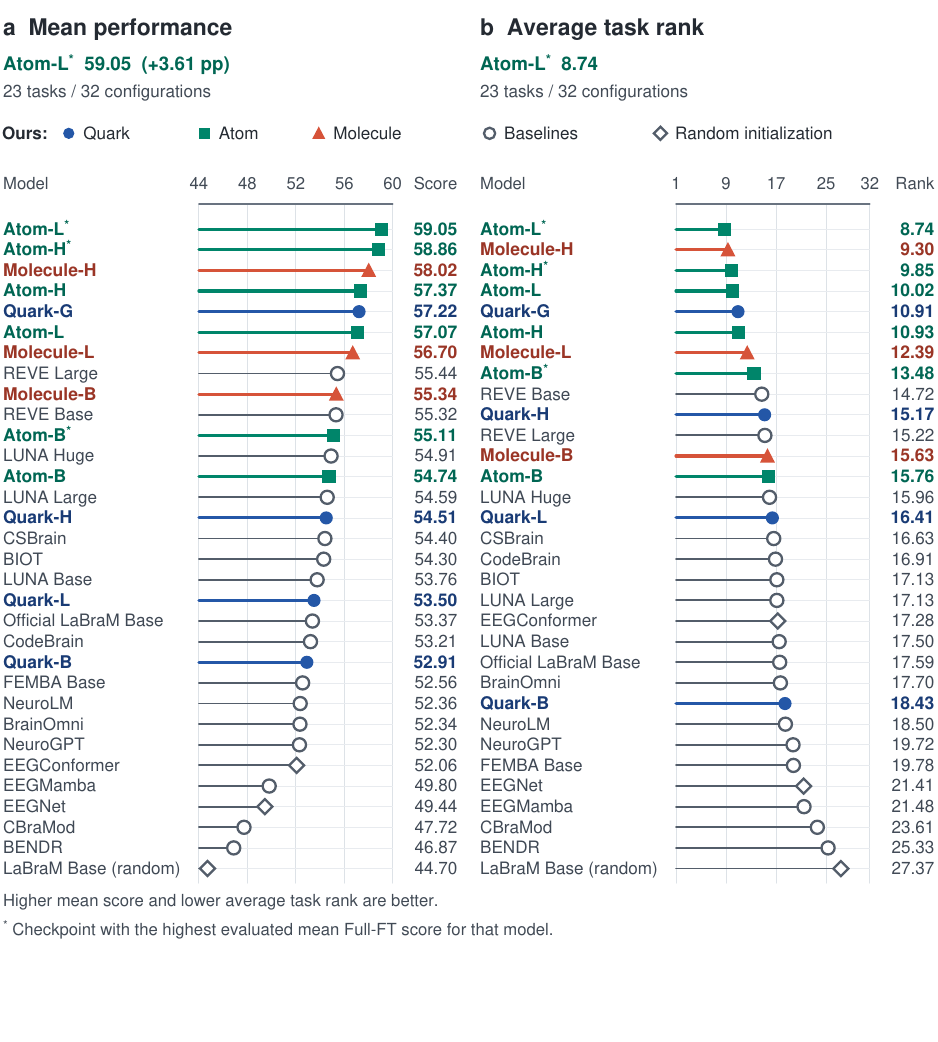}
\caption{Full-FT ranked by mean balanced accuracy (left) and average task rank (right) across 32 configurations and 23 tasks. Stars mark checkpoints selected by mean Full-FT score: Atom-B$^*$ (ep19), Atom-L$^*$ (ep29), and Atom-H$^*$ (ep19). Ties receive average ranks. The strongest baseline by mean score is REVE Large~\cite{elouahidi2025reve}.}
\label{fig:model-comparison}
\end{reportfigure}

\subsection{Linear probing}
\label{sec:linear-probing}

The gains carry over to frozen-feature transfer, with a different ordering among the leading encoders (Figure~\ref{fig:lp-rankings}). Quark-G leads LP at 54.33\% mean balanced accuracy, 1.24 percentage points above CodeBrain, the strongest baseline at 53.08\%~\cite{ma2026codebrain}. It also has the best average task rank at 8.63. \mbox{Atom-H$^*$} and \mbox{Atom-L$^*$} remain close in mean score at 54.25\% and 54.09\%. These checkpoints improve LP over their last evaluated counterparts, although they were selected using Full-FT scores.

\begin{reportfigure}
\centering
\includegraphics[width=\linewidth,trim=0 61 0 6,clip]{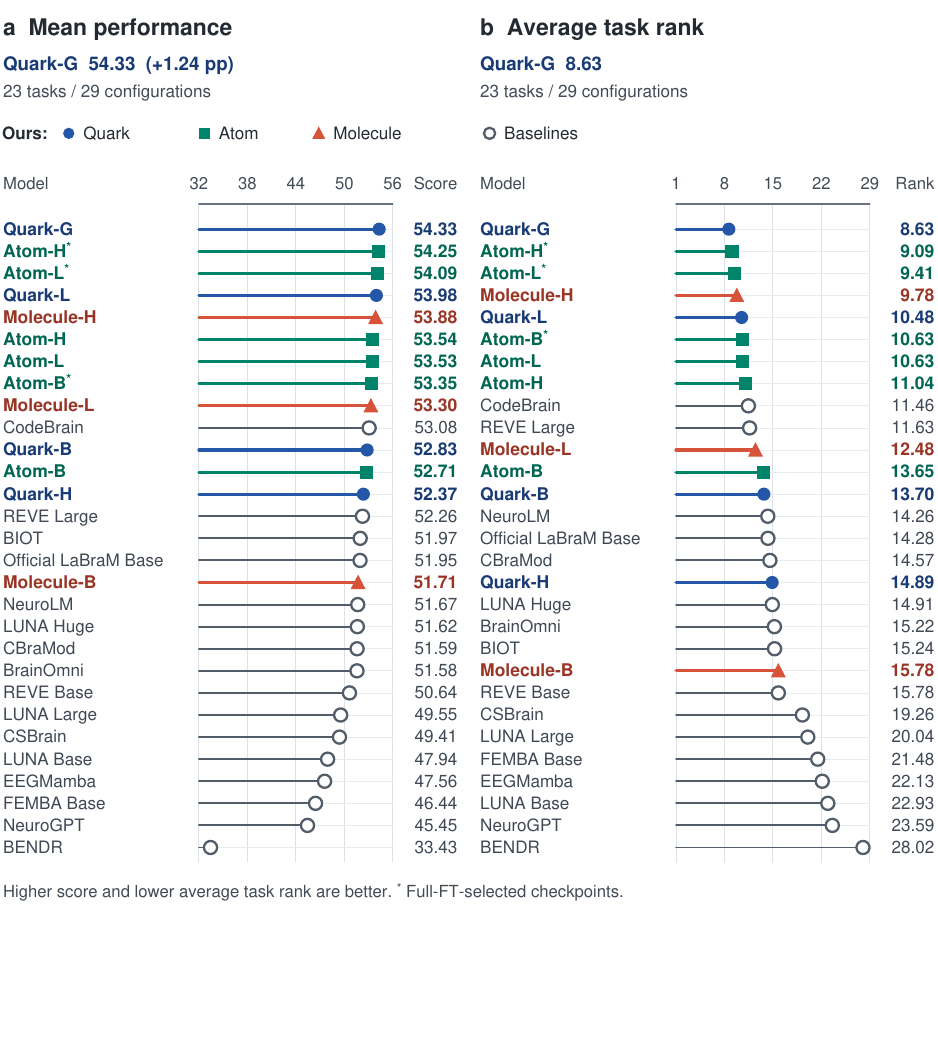}
\caption{LP ranked by mean balanced accuracy (left) and average task rank (right) across 29 configurations of 26 pretrained models and 23 tasks. Stars use the Full-FT-selected EEG checkpoints. Each task uses the higher three-seed mean of the available Pooled and No-pool readouts.}
\label{fig:lp-rankings}
\end{reportfigure}
Stronger LP performance does not necessarily imply stronger Full-FT performance. CodeBrain, for example, exceeds REVE Large in LP (53.08\% versus 52.26\%) and trails it in Full-FT (53.21\% versus 55.44\%). Across the 26 distinct pretrained encoders, 80 of 325 model pairs (24.6\%) reverse order between the two settings (Appendix~\ref{app:ft-lp-ordering}). Full-FT also reaches a higher observed performance ceiling on this benchmark: the best evaluated configuration scores 59.05\%, compared with 54.33\% under LP.

Average pooling can suppress task-relevant information by collapsing patch features across channels and time. No-pool avoids this aggregation by retaining the ordered patches for a larger linear classifier, raising mean balanced accuracy from 44.74\% to 49.61\% (+4.87 points) across 26 models evaluated on the same 22 tasks. We further hypothesize that larger, stronger encoders learn more general representations whose alignment with a specific downstream task is more demanding. These representations may require more substantial task-specific reorganization to realize their potential. Full-FT can make such adjustments by updating the encoder, whereas LP relies on a fixed representation.

\subsection{Task-level differences}
\label{sec:task-differences}
\label{sec:small-models}

\begin{reportfigure}
\centering
\includegraphics[width=0.95\linewidth,trim=0 5 0 4,clip]{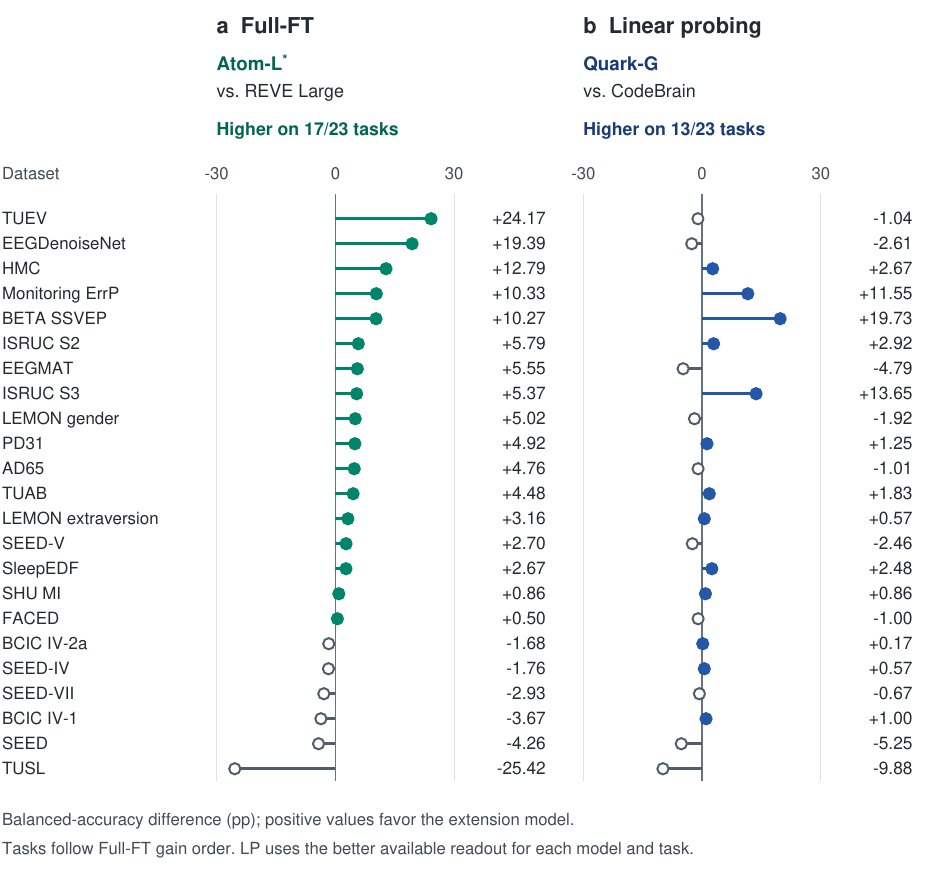}
\caption{Task-level score differences between the leading extension configuration and baseline by mean score in each setting. Positive values favor the extension. The Full-FT panel uses \mbox{Atom-L$^*$}; both panels cover 23 tasks.}
\label{fig:task-gains}
\end{reportfigure}

The individual task results show both the breadth and the concentration of the gains (Figure~\ref{fig:task-gains}). \mbox{Atom-L$^*$} exceeds REVE Large on 17 of 23 Full-FT tasks, with its largest gains on TUEV (+24.17 points), EEGDenoiseNet (+19.39), and HMC (+12.79). It trails on six tasks. For LP, Quark-G exceeds CodeBrain on 13 tasks and trails on ten. Its largest gains are on BETA SSVEP (+19.73 points), ISRUC S3 (+13.65), and Monitoring ErrP (+11.55), while TUSL and SEED remain weaknesses.

To assess the advantage over small models trained from scratch, we use the higher three-seed mean balanced accuracy of EEGNet and EEGConformer as the Full-FT reference for each task (Table~\ref{tab:checkpoint-scratch}). The released LaBraM Base exceeds it on 10 of 23 tasks. Among the 16 baseline EEG foundation models, REVE Base and Large lead with 12 wins each. Atom-L and Molecule-H each win on 16 tasks; Atom-L ties on one and trails on six, while Molecule-H trails on seven. Molecule-H retains all 12 wins of REVE Large and adds AD65, SEED-IV, HMC, and EEGDenoiseNet.

\subsection{Pretraining data, encoder size, and downstream performance}
\label{sec:resources}

\begin{reportfigure}
\centering
\includegraphics[page=1,width=0.92\linewidth,trim=0 16 0 6,clip]{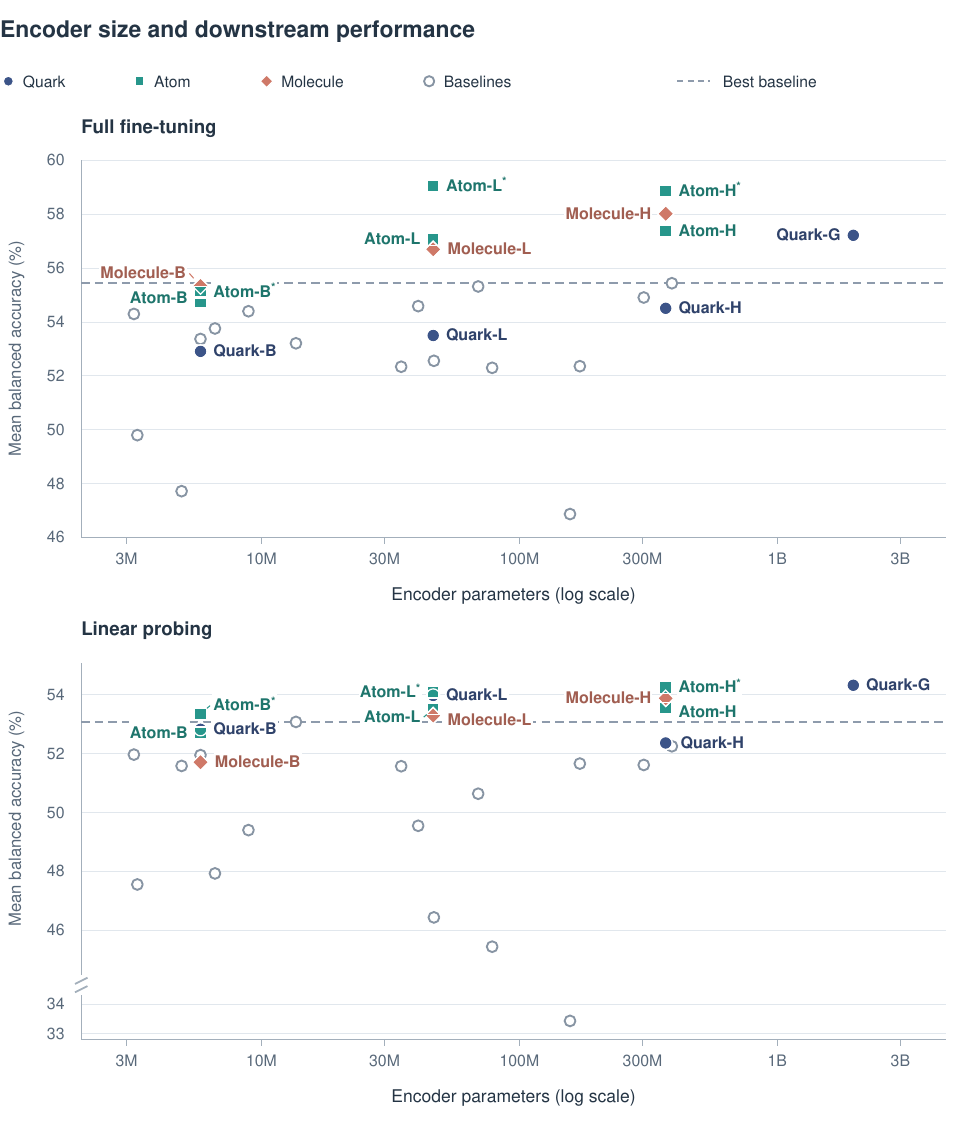}
\caption{Encoder size versus Full-FT (top) and LP (bottom) performance, including the three Full-FT-selected Atom references ($^*$). Dashed lines mark the highest baseline score; LP uses a broken vertical axis.}
\label{fig:resources-performance}
\end{reportfigure}

The model comparisons can also be viewed in terms of encoder size and pretraining hours (Figure~\ref{fig:resources-performance}). \mbox{Atom-L$^*$} reaches 59.05\% in Full-FT with 46.3M parameters and 49,628 hours, compared with 55.44\% for REVE Large at 390M parameters and 61,415 hours. The last evaluated Atom-L checkpoint reaches 57.07\%. Both comparisons favor Atom-L with a smaller encoder and fewer pretraining hours.

LP places the leading models much closer together despite large differences in size. Quark-G reaches 54.33\% with 1.97B parameters and 25,067 pretraining hours; Quark-L reaches 53.98\% with 46.3M parameters on the same collection. CodeBrain reaches 53.08\% with 13.6M parameters and approximately 9,246 hours. Quark-L thus approaches Quark-G's LP score with a much smaller encoder, while CodeBrain remains competitive with a smaller model and corpus.

\begin{reportfigure}
\ContinuedFloat
\centering
\includegraphics[page=2,width=0.92\linewidth,trim=0 16 0 6,clip]{resources_vs_performance.pdf}
\caption[]{Pretraining duration versus Full-FT (top) and LP (bottom) performance. Published durations include estimates and lower bounds (EEG and MEG for BrainOmni). NR: not reported; Atom-L/H labels two nearly coincident LP points.}
\end{reportfigure}

\section{Discussion}
\label{sec:observations}

Scaling encoder capacity produces consistent Full-FT gains across the three data versions. The benefit of additional pretraining hours is more dependent on capacity, with fraction sweeps suggesting a saturation region for each model size. Training trajectories reveal a further limit: continued improvement on the pretraining objective can accompany declining downstream performance.

\subsection{Model size}

Increasing encoder capacity improves Full-FT within each pretraining collection (Figure~\ref{fig:scale}). At the last evaluated checkpoints, Full-FT increases from Base to Huge by 1.60 points for Quark, 2.63 for Atom, and 2.68 for Molecule. Quark-G extends this trend to 57.22\%, 2.71 points above Quark-H.

\begin{reportfigure}
\centering
\includegraphics[width=\linewidth]{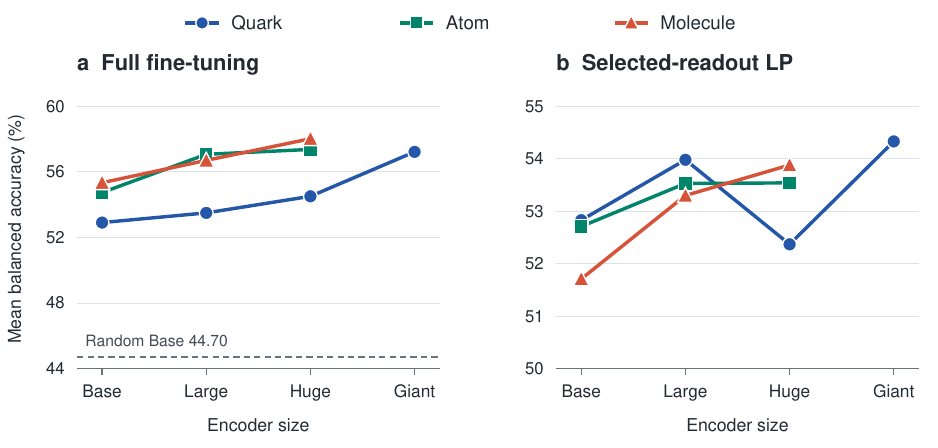}
\caption{Performance by data version and encoder size at the last evaluated checkpoints. Full-FT improves with capacity within each version; selected-readout LP varies less consistently. The dashed line marks the randomly initialized Base Full-FT result.}
\label{fig:scale}
\end{reportfigure}

The same size comparisons give less consistent gains under LP. Quark falls from 53.98\% at Large to 52.37\% at Huge, Atom-L and Atom-H are nearly tied at 53.53\% and 53.54\%, and Molecule improves from Base to Huge. Capacity therefore benefits Full-FT more consistently than frozen-feature transfer in these runs. This pattern is consistent with the distinction in Section~\ref{sec:linear-probing}: increasing capacity can improve performance after task-specific adaptation without a corresponding gain in the linear accessibility of frozen features.

\subsection{Pretraining data duration}

At a fixed encoder size, the data versions provide a first comparison of pretraining corpus duration. Atom and Molecule exceed Quark in Full-FT at each shared size. Molecule improves over Atom at Base and Huge, and falls slightly below it at Large (56.70\% versus 57.07\%).

At Base size, random initialization and the released LaBraM provide two further references for these recipes. Quark-B, Atom-B, and Molecule-B reach 52.91\%, 54.74\%, and 55.34\% Full-FT, all above the randomly initialized Base encoder at 44.70\%. Against the released LaBraM Base at 53.37\%, Atom-B and Molecule-B improve by 1.37 and 1.97 points, while Quark-B is 0.46 points lower.

The reduced-data Atom runs also outperform Quark at the same encoder sizes. With 10\%, 20\%, and 50\% of Atom, corresponding to approximately 4,963, 9,926, and 24,814 hours, Atom-B reaches 54.54\%, 53.40\%, and 56.00\% Full-FT, while Atom-L reaches 54.98\%, 55.26\%, and 57.15\%. All six scores exceed their Quark counterparts trained on 25,067 hours (52.91\% for Base and 53.50\% for Large). Atom-L approaches REVE with 10--20\% of Atom; at 50\%, both Atom sizes exceed REVE Base and Large (55.32\% and 55.44\%), pretrained on 61,415 hours~\cite{elouahidi2025reve}.

A possible explanation is the breadth of Atom's pretraining sources. Its pool spans 521 datasets, compared with 9 for Quark and 92 for REVE~\cite{elouahidi2025reve}. Sampling windows across this broader pool may expose the encoder to more varied tasks and recording conditions, improving transfer even with fewer recording hours. This interpretation is consistent with OmniEEG-Bench, which reports an association between more pretraining datasets and stronger downstream rankings under LP~\cite{lu2026omnieegbench}.

Within each collection, the fraction sweeps also show that gains from additional hours can level off or reverse (Figure~\ref{fig:data-fractions}). Molecule-B is strongest at 20\% data, approximately 16,000 hours: its Full-FT score falls from 56.35\% there to 55.34\% on the full collection, and LP falls from 53.09\% to 51.71\%. Atom-B peaks later, at 50\% data (approximately 24,800 hours), before both scores decline. Atom-L's Full-FT is nearly unchanged between 50\% and 100\% data, approximately 24,800--49,600 hours (57.15\% versus 57.07\%), although LP improves over this interval.

\begin{reportfigure}
\centering
\includegraphics[width=\linewidth]{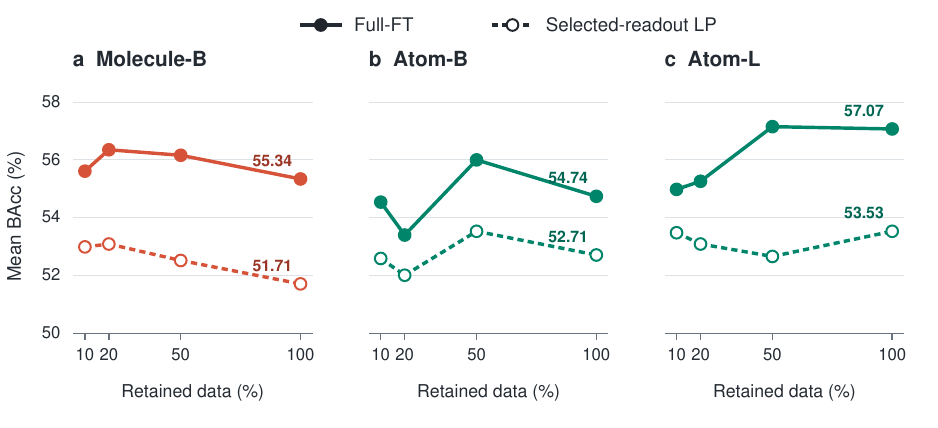}
\caption{Full-FT and selected-readout LP across retained data fractions. Each series uses a 50-epoch schedule, so the fraction also changes the number of training updates. Full-data references use the same checkpoints as the main comparisons.}
\label{fig:data-fractions}
\end{reportfigure}

These sweeps use 50 epochs, so increasing the data fraction also increases optimizer updates. Hours are approximated by multiplying each collection's duration by the retained fraction. The curves therefore describe the combined effect of corpus size and training exposure under this schedule. Their turning regions motivate a closer look at how the useful amount of data depends on encoder capacity.

\subsection{Capacity-dependent data saturation}
\label{sec:data-saturation}

At the sampled fractions, Base shows a decline where Large maintains similar Full-FT performance. This suggests that a fixed-size encoder has a saturation region and that larger encoders can benefit from more pretraining data. The broad spacing between fractions leaves the peak durations uncertain. A shared Full-FT response provides an estimate between the observed points, with duration normalized as $u=H/H_*(P)$:
\begin{equation}
\widehat S_i(H)=A_i+B\,\frac{2u}{1+u^2},\qquad
u=\frac{H}{H_*(P_i)},\qquad
H_*(P)=K\left(\frac{P}{5.8\times10^6}\right)^\eta.
\label{eq:ft-scaling-response}
\end{equation}
Here $H$ is corpus duration without epoch repetition, $P$ counts encoder parameters, and $i$ identifies a model--collection series. The three series share amplitude $B\geq0$ and response shape, with separate levels $A_i$. The response rises below $u=1$ and falls above it, balancing two fixed power terms in $u$ and $u^{-1}$.

Fitting the 12 Full-FT observations gives the following rounded peak-duration relation and its inverse:
\begin{equation}
\begin{aligned}
H_*(P)&\approx 22{,}100
\left(\frac{P}{5.8\times10^{6}}\right)^{0.292}\ \mathrm{hours},\\[3pt]
P_{\mathrm{ref}}(H)&\approx 5.8\times10^{6}
\left(\frac{H}{22{,}100\ \mathrm{hours}}\right)^{1/0.292}.
\end{aligned}
\label{eq:data-saturation}
\end{equation}
The fit places the Base peak near 22,000 hours and the Large peak near 40,000 hours. This empirical scaling hypothesis links Base and Large; its inverse maps duration to encoder size along the same fitted peak relation. Atom-B's 20\% point receives weight 0.25, and the Large peak is restricted to 24,814--49,628 hours. Appendix~\ref{app:ft-duration-fit} gives the fitting assumptions and sensitivity checks.

\subsection{The gap between pretraining and downstream evaluation}

Lower pretraining loss does not imply stronger downstream performance. Full-FT can peak before the last evaluated checkpoint (Figure~\ref{fig:loss-transfer}). Atom-L illustrates this directly: between ep29 and ep49, its loss decreases from 6.214 to 6.079 while Full-FT falls from 59.05\% to 57.07\%. Molecule-L shows a similar loss of about two Full-FT points during continued loss reduction, with declines on 16 of 23 tasks. The deterioration therefore extends across much of its downstream task profile.

The gap also appears in frozen features. Between ep9 and ep49, Molecule-B's loss falls from 8.322 to 7.949, while selected-readout LP declines from 52.98\% to 51.71\%. The aligned FT and LP trajectories in Figure~\ref{fig:loss-transfer} show that this mismatch occurs both when the encoder is frozen and when it is fully fine-tuned.

\begin{reportfigure}
\includegraphics[width=\linewidth]{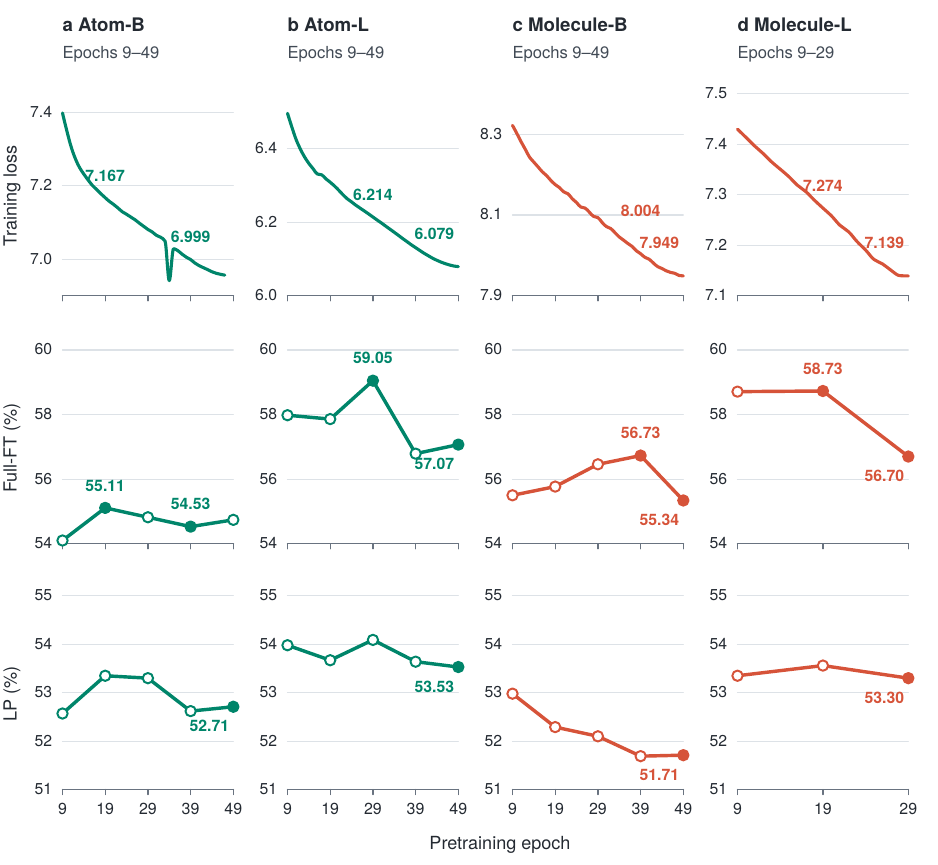}
\caption{Pretraining loss (top), Full-FT (middle), and selected-readout LP (bottom) along four pretraining runs. Atom-B's loss ends at ep47; its downstream results extend to ep49.}
\label{fig:loss-transfer}
\end{reportfigure}

Further optimization can produce lower pretraining loss and worse downstream performance. Training loss alone is therefore insufficient for choosing a training duration or a transfer checkpoint; both choices require downstream evaluation. Appendix~\ref{app:training-dynamics} provides the complete recorded trajectories.

\section{A Shared Language-Model Readout for EEG Tasks}
\label{sec:eeg-llm}

NeuroLM~\cite{jiang2025neurolm}, UniMind~\cite{lu2025unimind}, and GVG~\cite{pan2026gvg} explore unified EEG decoding across multiple tasks using language or multimodal language models. Inspired by these studies, we examine whether the gains from scaling LaBraM extend to a shared model trained jointly on all 23 tasks. This evaluation compares pretraining collections, EEG encoder capacities, and language-model sizes.

\subsection{Connecting frozen EEG features to Qwen3}

A shared Qwen3 readout~\cite{yang2025qwen3} classifies all 23 tasks from frozen EEG features. Task prompts specify the problem and candidate labels. Channel-averaged features are standardized with training-split statistics, pooled or interpolated to 32 tokens, and mapped to Qwen3's input dimension by a two-layer MLP. Inference selects the highest-scoring textual class candidate.

Joint training updates the projector and LoRA adapters~\cite{hu2022lora}, keeping the EEG encoder and Qwen3 base weights frozen. Training uses the same 40-window subject--class cap as the single-task evaluations. Mean validation balanced accuracy selects one shared checkpoint per seed for full-test evaluation. Appendix~\ref{app:eeg-llm} gives settings, splits, sample counts, and per-task scores with seed variability.

\subsection{Data, encoder, and language-model scale}

We pair ten EEG encoders with Qwen3-0.6B, 1.7B, 4B, 8B, 14B, and 32B: 60 configurations with three seeds each (Figure~\ref{fig:eeg-llm-results}).

\begin{reportfigure}
\includegraphics[width=\linewidth]{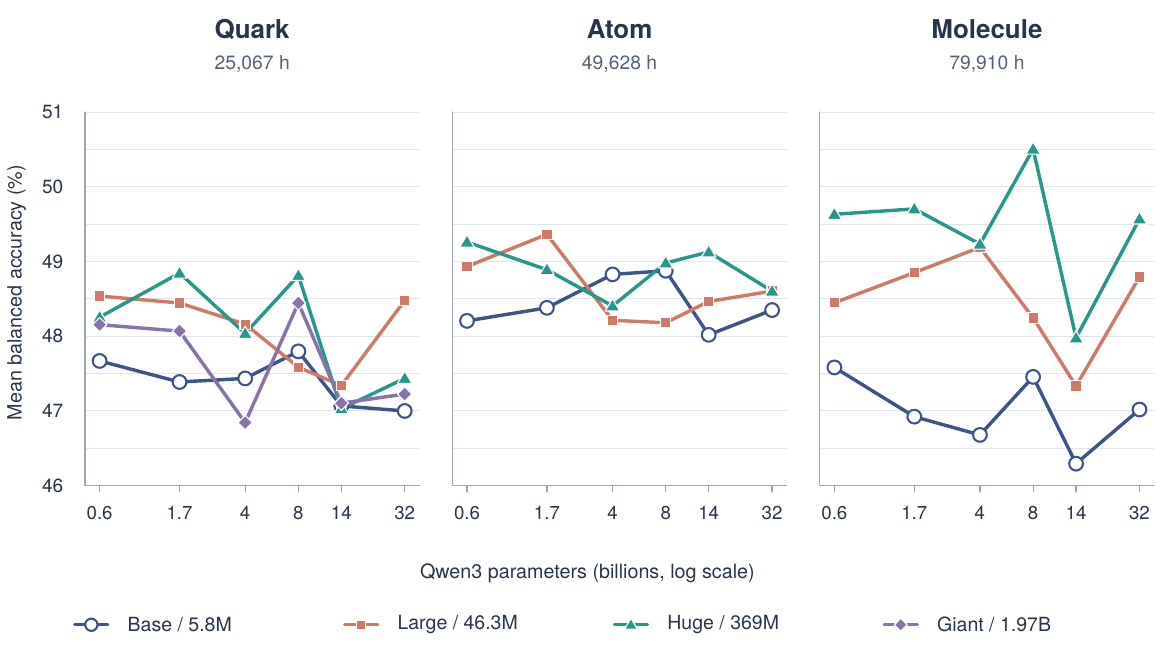}
\caption{Joint classification across 23 EEG tasks with ten encoders and six Qwen3 sizes. Panels compare pretraining collections; curves distinguish encoder sizes. Scores are three-seed means on a shared vertical scale.}
\label{fig:eeg-llm-results}
\end{reportfigure}

Pretraining data and EEG encoder capacity show complementary trends in joint classification. Moving from Quark's 25,067 hours to Atom's 49,628 hours improves performance in all 18 comparisons at matched EEG and Qwen3 sizes. With Molecule's 79,910 hours, further gains over Atom occur in five of six Huge comparisons and three of six Large comparisons, while Base performs worse throughout. The encoder comparisons reinforce this pattern: at a fixed pretraining collection and Qwen3 size, Huge (369M) outperforms Base (5.8M) in 16 of 18 comparisons. For Molecule, performance increases from Base through Large (46.3M) to Huge at all six Qwen3 sizes, with an average Base-to-Huge gain of 2.44 points.

Encoder orderings can change when moving from linear probing to joint EEG--LLM decoding. Quark-G (1.97B) scores higher than Quark-H under Pooled LP (51.11\% versus 48.05\%), yet exceeds it in only one of the six Qwen3 settings. Larger language-model readouts also give inconsistent gains: every EEG encoder reaches its highest measured mean with Qwen3-8B or smaller, and Qwen3-32B exceeds 0.6B for only two of the ten encoders. Molecule-H with Qwen3-8B achieves the highest mean, \mbox{$50.50\pm2.17\%$}.

The Pooled/No-pool results in Section~\ref{sec:linear-probing} offer a possible explanation for the limited return from a larger readout. Retaining individual patches improves LP over average pooling. Here, channel features are also averaged, although a temporal sequence of 32 tokens is retained. Qwen3 therefore lacks direct access to the individual channel tokens, and the frozen EEG encoder cannot adjust its features during joint training. Information retention and feature adaptation may thus limit how much additional language-model capacity helps.

The clearest gains in joint decoding come from combining larger EEG pretraining collections with greater encoder capacity. The LP comparisons suggest that preserving EEG structure and improving its adaptation to the shared readout are important for realizing these gains; increasing LLM parameters alone provides no consistent improvement.

\section{Conclusion}
\label{sec:status}
\label{sec:overlap-subsets}

Scaling pretraining data and encoder capacity while retaining LaBraM's core encoder and objective yields strong downstream representations. Across 23 tasks, \mbox{Atom-L$^*$} leads Full-FT at 59.05\% and Quark-G leads LP at 54.33\%; both also lead average task rank. Against the better of EEGNet and EEGConformer on each task, Atom-L and Molecule-H each win on 16 of 23 Full-FT tasks, compared with at most 12 for a baseline EEG foundation model.

The benefits of capacity depend on downstream adaptation. Increasing encoder size improves Full-FT more consistently than LP, and higher LP scores do not guarantee higher Full-FT performance. No-pool's gains over average pooling suggest that retaining patch structure helps make frozen features useful to downstream tasks. The FT/LP ordering differences further support evaluating both linear accessibility and task-specific adaptation when judging the value of a pretrained encoder.

Data breadth and duration also shape the gains from pretraining. The reduced-data Atom results suggest that source diversity may improve data efficiency, while the fraction sweeps indicate a saturation region at fixed encoder capacity. The Full-FT fit for Base and Large gives $H_*\propto P^{0.292}$ and places the peaks near 22,000 and 40,000 hours, respectively, providing an empirical hypothesis for matching data duration to capacity (Equation~\ref{eq:data-saturation}). Continued optimization introduces a further limit: Full-FT and LP can decline while pretraining loss keeps falling. Training duration and checkpoint selection therefore require downstream evaluation.

Joint EEG--LLM decoding extends this dependence on adaptation to a shared readout. Larger EEG collections and encoders show complementary gains, while increasing Qwen3 size provides no consistent improvement. All ten encoders reach their highest mean with 8B or smaller, and Molecule-H with Qwen3-8B leads at 50.50\%. The ordering changes between Pooled LP and joint decoding, together with the pooling analysis, suggest that information preservation and feature adaptation may limit the benefit of additional language-model capacity.

LaBraM's performance supports expanding an established EEG pretraining design as a route to stronger foundation models. Further progress is likely to come from matching data breadth and duration to encoder capacity, and improving how the resulting representations are preserved and adapted for downstream tasks.

\begingroup
\renewcommand{\bibfont}{\small}
\setlength{\bibsep}{2.5pt}
\let\reportthebibliography\thebibliography
\renewcommand{\thebibliography}[1]{\reportthebibliography{#1}\interlinepenalty=100}
\pdfbookmark[1]{References}{references}
\bibliographystyle{unsrtnat}
\bibliography{references,dataset_references}
\endgroup
\appendix
\section{Pretraining Datasets}
\label{app:pretraining-datasets}

Table~\ref{tab:pretraining-inventory} lists all 523 dataset entries, with links to publications and data records. Quark, Atom, and Molecule contain 9, 521, and 505 entries, respectively. Durations are the recording hours used in each collection; a dash denotes absence. Accessions joined by a plus sign form one entry.

Numbered citations identify publications in the References; Data links point to repository records. A dagger marks a related publication whose exact correspondence to the listed dataset remains unconfirmed; unresolved aliases are labeled explicitly. The accompanying CSV files contain full source metadata, recording hours, and training-window counts.

\begingroup
\fontsize{8.7}{10.2}\selectfont
\setlength{\tabcolsep}{2.5pt}
\renewcommand{\arraystretch}{1.08}
\setlength{\LTpre}{4pt}
\setlength{\LTpost}{4pt}
\begin{longtable}{@{}>{\raggedright\arraybackslash}p{0.32\textwidth}>{\raggedright\arraybackslash}p{0.32\textwidth}*{3}{>{\raggedleft\arraybackslash}p{0.10\textwidth}}@{}}
\caption{Pretraining datasets, source links, and recording hours used in each collection.}\label{tab:pretraining-inventory}\\
\toprule
Dataset or accession & Source & Quark (h) & Atom (h) & Molecule (h) \\
\midrule
\endfirsthead
\caption[]{Pretraining datasets, sources, and recording hours (continued).}\\
\toprule
Dataset or accession & Source & Quark (h) & Atom (h) & Molecule (h) \\
\midrule
\endhead
\midrule
\multicolumn{5}{r@{}}{\emph{Continued on the next page.}}\\
\endfoot
\bottomrule
\endlastfoot
\multicolumn{5}{@{}l}{\rule{0pt}{14pt}\textbf{Named datasets and other repositories}}\\*
Alljoined & \href{https://huggingface.co/datasets/Alljoined/Alljoined-1.6M}{Data}; \cite{datasetf59c61126677} & 370.96 & -- & 461.03 \\
BCI-\allowbreak{}IV-\allowbreak{}2a & \href{https://www.bbci.de/competition/iv/}{Data}; \cite{dataset31613ebabb7c} & -- & 7.82 & 18.23 \\
BCIC2020\_\allowbreak{}3 & \href{https://osf.io/pq7vb/}{Data}; \cite{datasete2d21212d5dd} & -- & 5.00 & 5.00 \\
CHB-\allowbreak{}MIT & \href{https://physionet.org/content/chbmit/1.0.0/}{Data}; \cite{dataset9411f28d9da7} & -- & 991.45 & 1,963.72 \\
EEGMAT & \href{https://physionet.org/content/eegmat/1.0.0/}{Data}; \cite{dataset8e30c5b3be7e} & -- & 1.44 & 2.37 \\
EEGMMIDB & \href{https://physionet.org/content/eegmmidb/1.0.0/}{Data}; \cite{dataset9fd56b6bf574} & -- & 16.14 & 22.52 \\
EEGWorkload & \href{https://physionet.org/content/eegmat/1.0.0/}{Data 1}; \href{https://arxiv.org/html/2409.00101}{Data 2}; \cite{dataset8e30c5b3be7e}\raisebox{0.4ex}{\fontsize{6.5}{7}\selectfont †} & -- & 1.68 & -- \\
\end{longtable}
\endgroup

\section{Pretraining Data Details}
\label{app:additional-settings}

\subsection{Source composition}

Table~\ref{tab:data-composition} gives the source breakdown behind Figure~\ref{fig:data-composition}. Categories summarize the predominant recording setting or task. Quark uses sampling weight 1.0 for TUH, Alljoined, and THINGS-EEG2; 0.1 for HMC; and 0.3 for the self-collected dataset and the four SEED datasets. Their recording hours appear in Table~\ref{tab:pretraining-named-sources}.

\begingroup
\small
\setlength{\LTpre}{6pt}
\setlength{\LTpost}{8pt}
\begin{longtable}{@{}lrrrr@{}}
\caption{Source composition of Atom and Molecule.}\label{tab:data-composition}\\
\toprule
Category & Atom (h) & Share & Molecule (h) & Share \\
\midrule
\endfirsthead
\caption[]{Source composition of Atom and Molecule (continued).}\\
\toprule
Category & Atom (h) & Share & Molecule (h) & Share \\
\midrule
\endhead
\midrule
\multicolumn{5}{r@{}}{\emph{Continued on the next page.}}\\
\endfoot
\bottomrule
\endlastfoot
Routine clinical (TUH) & 22,554 & 45.4\% & 22,158 & 27.7\% \\
Cognitive / ERP & 6,358 & 12.8\% & 11,420 & 14.3\% \\
Unclassified / mixed & 6,235 & 12.6\% & 4,076 & 5.1\% \\
Sleep PSG & 4,704 & 9.5\% & 4,551 & 5.7\% \\
Auditory / speech & 1,954 & 3.9\% & 1,459 & 1.8\% \\
ICU (I-CARE) & 1,657 & 3.3\% & 28,583 & 35.8\% \\
Motor imagery & 1,333 & 2.7\% & 1,429 & 1.8\% \\
Epilepsy, excluding TUH & 1,229 & 2.5\% & 2,309 & 2.9\% \\
Other clinical cohorts & 909 & 1.8\% & 822 & 1.0\% \\
Emotion & 793 & 1.6\% & 726 & 0.9\% \\
Resting state & 790 & 1.6\% & 699 & 0.9\% \\
SSVEP / P300 and related BCI & 635 & 1.3\% & 588 & 0.7\% \\
Visual image evoked & 477 & 1.0\% & 1,089 & 1.4\% \\
\end{longtable}
\tabnote{Hours and percentages are rounded independently.}
\endgroup

\subsection{Windowing, sampling, and padding}

For channel count $C$, Molecule uses $\min(16,\max(4,\lfloor256/C\rfloor))$ seconds per window, except for the one-second Alljoined and THINGS-EEG2 trials. Above 64 channels, the four-second minimum permits more than 256 channel-by-time tokens. Quark truncates longer samples to 16 one-second patches.

At retained fraction $f$, the loader keeps every $\operatorname{round}(1/f)$-th window within each source/view stream. Selected windows recur across epochs.

The node-local shard cache defaults to 80\% of disk capacity; the loader supports persistent workers, four-batch prefetching, and pinned host memory.

The student uses patch-validity weights and excludes fully padded patches as described in Section~\ref{sec:model}. The tokenizer's reconstruction loss includes zero-padded patches.

\subsection{Molecule preparation settings}
\label{app:molecule-preparation}

Table~\ref{tab:molecule-preparation} specifies the signal preparation and routing rules used to rebuild Molecule. Quark and Atom retain their earlier preparation pipelines. Atom and Quark use input scale 0.01; Molecule uses 1.0 after recording- or trial-level normalization. Window construction requires finite values, identified channels, and explicit units and padding metadata.

During training, the Atom and Molecule input paths sanitize non-finite values and clip tokenizer inputs to $\pm20$. For Molecule, the same sanitation is applied before frozen-target extraction and encoder processing.

\begingroup
\fontsize{9}{10.8}\selectfont
\renewcommand{\arraystretch}{1.08}
\setlength{\LTpre}{6pt}
\setlength{\LTpost}{8pt}
\begin{longtable}{@{}>{\raggedright\arraybackslash}p{0.23\linewidth} >{\raggedright\arraybackslash}p{0.73\linewidth}@{}}
\caption{Signal preparation and window routing for Molecule.}\label{tab:molecule-preparation}\\
\toprule
Property & Setting \\
\midrule
\endfirsthead
\caption[]{Signal preparation and window routing for Molecule (continued).}\\
\toprule
Property & Setting \\
\midrule
\endhead
\midrule
\multicolumn{2}{r@{}}{\emph{Continued on the next page.}}\\
\endfoot
\bottomrule
\endlastfoot
\multicolumn{2}{@{}l}{\textbf{Signal and window construction}} \\
Sampling and filtering & 200 Hz; band-pass 0.5--$\min(99.5,\,0.99f_{\mathrm{Nyquist}})$ Hz; no notch filter. \\
Bad channels & Remove exactly constant channels and channels with non-finite fraction $\geq 1\%$. \\
Finite segments & Split the remaining signal into contiguous finite runs of at least 200 samples; no non-finite imputation. \\
Normalization & Per-channel z-score within each recording or trial, with $\epsilon=10^{-7}$; clip to $\pm15$; store in float16; training input scale 1.0. \\
Window duration & $\min(16,\max(4,\lfloor256/C\rfloor))$ s for $C$ mapped channels. Alljoined and THINGS-EEG2 use 1 s trials. Windows do not overlap or cross recording/trial boundaries. \\
Recording tails & Retain tails of at least 1 s; zero-pad and record sample masks and valid-patch fractions. \\
\midrule
\multicolumn{2}{@{}l}{\textbf{Quality flags and sampling}} \\
Mild quality flag & Any channel with $|\mu|>5$ or $\sigma<0.05$ or $\sigma>5$; any clipping or padding; any removed bad channel or reader fallback. \\
Severe quality flag & Any channel with $|\mu|>10$ or $\sigma<0.005$ or $\sigma>10$; clipped fraction $>5\%$; padded fraction $\geq50\%$; or all channels bad/flat. \\
Exclusion priority & Subject exclusions, non-finite/decoding failures, and severe-quality flags precede auxiliary routing and receive zero training weight. \\
Auxiliary factors & Mild quality flag: 0.5. One to seven channels: 0.25, or 0.125 with a mild flag. Main view: 1.0. \\
Sampling factors & Multiply source, channel, quality, and provenance factors. Approved processed fallbacks have provenance factor 0.5; raw recordings and complete trial caches use 1.0. \\
\end{longtable}
\tabnote{Signal statistics use valid samples in normalized units; $\mu$ and $\sigma$ denote a channel's window mean and standard deviation. A severe flag overrides a mild flag. Sampling factors determine window acceptance probabilities.}
\endgroup

\subsection{Tokenizer settings}

The three versions share the tokenizer architecture. Each uses independently trained weights for frozen target generation during encoder pretraining.

\begin{reporttable}
\centering
\small
\caption{Tokenizer training configurations.}
\label{tab:tokenizers}
\begin{tabular}{@{}lrrrcr@{}}
\toprule
Version & GPUs & Global batch & Peak LR & Warmup & Checkpoint \\
\midrule
Quark & 32 & 2,048 & $8\times10^{-4}$ & 10 & ep39 \\
Atom & 8 & 2,048 & $2.4\times10^{-4}$ & 15 & ep16 \\
Molecule & 32 & 6,144 & $2.4\times10^{-4}$ & 15 & ep39 \\
\bottomrule
\end{tabular}
\tabnote{Learning rates include batch-size scaling. Warmup is in epochs; planned schedules are 100, 40, and 40 epochs.}
\end{reporttable}

\section{Downstream Datasets and Evaluation Settings}
\label{app:tasks}
\label{app:evaluation-settings}

Task inputs are sampled at 200 Hz and retain auxiliary channels where present. TUEV files have 24--32 channels; the table describes the inspected 32-channel input. Extension models use sample/channel z-scoring and clipping to $\pm15$; external encoders retain their model-specific input scaling. For inputs longer than 16 seconds, such as 30-second sleep windows, we interpolate the extension encoders' temporal embeddings while retaining one-second patches.

\begingroup
\begingroup
\fontsize{9}{10}\selectfont
\setlength{\tabcolsep}{3pt}
\renewcommand{\arraystretch}{1.0}
\setlength{\LTpre}{6pt}
\setlength{\LTpost}{8pt}
\begin{longtable}{@{}llrrrrrrr@{}}
\caption{Downstream datasets, input configurations, and Seed42 window counts.}\label{tab:tasks}\\
\toprule
\multicolumn{6}{c}{Dataset and input configuration} & \multicolumn{3}{c}{Seed42 windows} \\
\cmidrule(lr){1-6}\cmidrule(l){7-9}
Dataset & Category & Classes & Units & Ch. & Window (s) & Train & Validation & Test \\
\midrule
\endfirsthead
\caption[]{Downstream datasets, input configurations, and Seed42 window counts (continued).}\\
\toprule
\multicolumn{6}{c}{Dataset and input configuration} & \multicolumn{3}{c}{Seed42 windows} \\
\cmidrule(lr){1-6}\cmidrule(l){7-9}
Dataset & Category & Classes & Units & Ch. & Window (s) & Train & Validation & Test \\
\midrule
\endhead
\midrule
\multicolumn{9}{r@{}}{\emph{Continued on the next page.}}\\
\endfoot
\bottomrule
\endlastfoot
AD65~\citep{dataset647b0bdbb1e9} & Neurodeg. & 3 & 88 & 19 & 10 & 2,790 & 816 & 697 \\
PD31~\citep{datasetacff5002cc78,dataset0b04146707ad} & Neurodeg. & 2 & 31 & 40 & 1 & 1,000 & 773 & 581 \\
BCIC IV-1~\citep{dataset44bf19108281,dataset31613ebabb7c} & Motor imag. & 2 & 7 & 59 & 4 & 400 & 200 & 200 \\
BCIC IV-2a~\citep{dataset31613ebabb7c} & Motor imag. & 4 & 9 & 22 & 4 & 1,120 & 576 & 576 \\
SHU MI~\citep{datasetf71ad7f65b6d} & Motor imag. & 2 & 25 & 32 & 4 & 1,600 & 977 & 1,455 \\
BETA SSVEP~\citep{dataset92af3ba4541c} & SSVEP & 40 & 70 & 64 & 1 & 8,960 & 1,120 & 1,120 \\
Monitoring ErrP~\citep{dataset7fee03c5b499} & ERP & 2 & 6 & 64 & 1 & 320 & 1,080 & 1,044 \\
EEGMAT~\citep{dataset8e30c5b3be7e} & Cognitive & 2 & 36 & 21 & 5 & 1,371 & 181 & 144 \\
LEMON gender~\citep{datasetcbc317708c6e} & Trait & 2 & 203 & 59 & 1 & 6,480 & 18,828 & 19,734 \\
LEMON extraversion~\citep{datasetcbc317708c6e} & Trait & 2 & 203 & 59 & 1 & 6,480 & 18,828 & 19,734 \\
FACED~\citep{dataset84a6347989cf} & Emotion & 9 & 123 & 32 & 10 & 8,232 & 1,008 & 1,092 \\
SEED~\citep{dataset0678103fbd8b} & Emotion & 3 & 15 & 62 & 10 & 1,440 & 1,986 & 993 \\
SEED-IV~\citep{dataset74f7ed3a6339} & Emotion & 4 & 15 & 62 & 4 & 1,920 & 5,010 & 2,505 \\
SEED-V~\citep{dataset099fab2a0498} & Emotion & 5 & 16 & 62 & 1 & 2,600 & 14,718 & 7,359 \\
SEED-VII~\citep{datasetd169eca259ba} & Emotion & 7 & 20 & 62 & 10 & 4,480 & 2,734 & 2,734 \\
ISRUC S2~\citep{datasetc1777412f8f9} & Sleep & 5 & 8 & 6 & 30 & 1,200 & 1,975 & 1,720 \\
ISRUC S3~\citep{datasetc1777412f8f9} & Sleep & 5 & 10 & 6 & 30 & 1,600 & 792 & 996 \\
SleepEDF~\citep{dataseted441eb14280} & Sleep & 5 & 153$^{a}$ & 2 & 30 & 22,790 & 40,989 & 43,334 \\
HMC~\citep{dataset8c058cac92db} & Sleep & 5 & 151 & 8 & 30 & 23,691 & 13,038 & 13,838 \\
TUAB~\citep{dataset9f715fb83759} & Clin. screen & 2 & 2,328 & 23 & 10 & 76,000 & 42,352 & 41,962 \\
TUEV~\citep{dataset4ddb45ef1162} & Clin. event & 6 & 370 & 32 & 5 & 16,409 & 13,206 & 12,908 \\
TUSL~\citep{datasete16adebcf4e5} & Clin. event & 3 & 30 & 23 & 10 & 201 & 28 & 16 \\
EEGDenoiseNet~\citep{dataset7ba4ba34c1c9} & Artifact & 2 & 1$^{b}$ & 1 & 2 & 80 & 1,351 & 1,352 \\
\end{longtable}
\endgroup
\tabnote{Neurodeg.: neurodegenerative disease; Motor imag.: motor imagery; Clin. screen: clinical screening; Clin. event: clinical events. Units are subjects, except $^{a}$153 nights in SleepEDF and $^{b}$one pseudo-subject in EEGDenoiseNet. Training is capped at 40 samples per subject/class unit; validation and test retain all windows. Other seeds use different splits.}
\endgroup

Seeds 42, 10, and 5 control splitting and downstream randomness. Splits target 80/10/10 by registered subject identifier. SleepEDF uses night identifiers as split units, allowing person-level overlap; EEGDenoiseNet uses a within-subject sample split. The training cap applies to Full-FT, LP, and joint EEG--LLM evaluation.

\subsection{Optimization and readouts}

Full-FT uses a 50-epoch schedule to train the encoder and head with AdamW, class-weighted loss, and label smoothing. The search pairs learning rates $\{10^{-4},3\times10^{-4},10^{-3}\}$ with layer-wise decay $\{0.75,1.0\}$. LP trains a linear head for 30 epochs at constant learning rates $\{10^{-4},5\times10^{-4},10^{-3}\}$ and weight decay $\{0,0.05,0.1\}$. Pooled averages patch features; No-pool flattens them in channel/time order. LP candidates and replication fits reuse memory-mapped frozen features within the same task and split; Full-FT fits each candidate independently.

Within each fit, validation scores select the epoch. Seed42 test scores select hyperparameters for replication with seeds 10 and 5. Combined LP results use the higher three-seed test mean of the available readouts, with Pooled used on ties and for SleepEDF.

BENDR uses its stable Full-FT grid except for BCIC IV-2a. BIOT's released weights include supervised TUAB and TUEV training~\cite{yang2023biot}. Its BCIC IV-2a No-pool result uses the fixed 18-channel coordinate mapping. On EEGDenoiseNet, BIOT receives only the native Cz channel. FEMBA Base uses a Cz-only montage: C3--Cz is $-\mathrm{Cz}$, Cz--C4 is $+\mathrm{Cz}$, and the other 20 bipolar channels are zero. This task adaptation differs from FEMBA's published input configuration~\cite{tegon2025femba}.

\section{Results after Excluding Known Overlapping Sources}
\label{app:overlap}

The 12-task subset contains BCIC IV-1, SHU MI, BETA SSVEP, Monitoring ErrP, LEMON gender, LEMON extraversion, ISRUC S2, SleepEDF, TUAB, TUEV, TUSL, and EEGDenoiseNet. The nine-task subset additionally removes TUAB, TUEV, and TUSL. These exclusions address identified or potential overlap; recording-level separation remains unverified for some retained task--model pairs.

\begingroup
\small
\begin{longtable}{@{}lrrrr@{}}
\caption{Results after excluding known overlapping sources.}\label{tab:overlap}\\
\toprule
Model & FT-12 & FT-9 & LP-12 & LP-9 \\
\midrule
\endfirsthead
\multicolumn{5}{l}{\small\itshape Continued from previous page}\\
\toprule
Model & FT-12 & FT-9 & LP-12 & LP-9 \\
\midrule
\endhead
\midrule
\multicolumn{5}{r}{\small Continued on next page}\\
\endfoot
\bottomrule
\endlastfoot
Quark-B & 60.86 & 58.53 & 58.07 & 55.85 \\
Quark-L & 60.82 & 58.65 & 60.28 & 58.00 \\
Quark-H & 61.00 & 57.22 & 59.28 & 57.44 \\
Quark-G & 62.21 & 59.22 & 60.75 & 59.05 \\
Atom-B & 60.63 & 57.69 & 58.38 & 55.35 \\
Atom-L & 63.75 & 60.36 & 59.41 & 57.26 \\
Atom-H & 63.05 & 59.79 & 59.32 & 57.34 \\
Molecule-B & 60.86 & 58.46 & 58.27 & 55.47 \\
Molecule-L & 62.20 & 59.53 & 60.15 & 57.35 \\
Molecule-H & 63.82 & 61.03 & 59.62 & 57.44 \\
REVE Base & 61.01 & 57.88 & 56.16 & 53.62 \\
REVE Large & 59.99 & 55.94 & 57.82 & 54.94 \\
LUNA Base & 58.72 & 55.81 & 54.63 & 52.25 \\
LUNA Large & 59.95 & 56.78 & 56.68 & 53.36 \\
LUNA Huge & 59.00 & 55.14 & 58.13 & 55.48 \\
CSBrain & 62.07 & 57.88 & 58.07 & 54.29 \\
Official LaBraM Base & 58.47 & 54.80 & 58.13 & 55.39 \\
CodeBrain & 58.15 & 54.39 & 58.62 & 55.20 \\
NeuroLM & 58.27 & 53.23 & 57.26 & 52.90 \\
BrainOmni & 57.60 & 53.68 & 57.47 & 55.49 \\
NeuroGPT & 57.74 & 55.86 & 53.57 & 52.00 \\
EEGMamba & 56.46 & 52.61 & 54.87 & 51.52 \\
CBraMod & 51.97 & 50.67 & 57.29 & 54.82 \\
BENDR & 52.91 & 52.02 & 38.91 & 39.69 \\
BIOT & 59.92 & 54.57 & 59.39 & 54.02 \\
FEMBA Base & 57.27 & 55.53 & 51.28 & 48.62 \\
EEGNet & 52.28 & 51.88 & \na & \na \\
EEGConformer & 56.08 & 56.99 & \na & \na \\
LaBraM Base, random & 50.31 & 48.64 & \na & \na \\
\end{longtable}
\endgroup
\tabnote{Extension models use the same checkpoints as Figure~\ref{fig:model-comparison}. LP-12 and LP-9 use selected-readout LP. A dash indicates an unavailable result.}

\section{Linear-Probing Readout Comparison}
\label{app:lp-readouts}

Table~\ref{tab:lp-paired-results} compares Pooled and No-pool on matched tasks at the last evaluated full-data checkpoints. All 26 models cover the same 22 paired tasks; SleepEDF has only Pooled results. The source files provide the available three-seed results, the Atom-H TUAB trajectory mean, and the combined LP rankings.

\begingroup
\small
\setlength{\tabcolsep}{8pt}
\renewcommand{\arraystretch}{1.05}
\setlength{\LTpre}{6pt}
\setlength{\LTpost}{8pt}
\begin{longtable}{lrrrr}
\caption{Matched comparison of LP readouts. Each model is compared on the same tasks under both readouts. Values are mean balanced accuracy (\%); $\Delta$ is No-pool minus Pooled in percentage points.}\label{tab:lp-paired-results}\\
\toprule
Model & Paired tasks & Pooled & No-pool & $\Delta$ \\
\midrule
\endfirsthead
\caption[]{Matched comparison of LP readouts. Each model is compared on the same tasks under both readouts. Values are mean balanced accuracy (\%); $\Delta$ is No-pool minus Pooled in percentage points (continued).}\\
\toprule
Model & Paired tasks & Pooled & No-pool & $\Delta$ \\
\midrule
\endhead
\midrule
\multicolumn{5}{r@{}}{\emph{Continued on the next page.}}\\
\endfoot
\bottomrule
\endlastfoot
\textbf{Quark-B} & 22 & 43.68 & 52.09 & +8.40 \\
\textbf{Quark-L} & 22 & 46.56 & 53.21 & +6.65 \\
\textbf{Quark-H} & 22 & 47.05 & 51.17 & +4.12 \\
\textbf{Quark-G} & 22 & 50.19 & 52.44 & +2.25 \\
\textbf{Atom-B} & 22 & 47.53 & 51.33 & +3.79 \\
\textbf{Atom-L} & 22 & 47.30 & 52.56 & +5.26 \\
\textbf{Atom-H} & 22 & 48.16 & 52.34 & +4.18 \\
\textbf{Molecule-B} & 22 & 45.94 & 50.58 & +4.64 \\
\textbf{Molecule-L} & 22 & 45.52 & 52.58 & +7.06 \\
\textbf{Molecule-H} & 22 & 48.66 & 52.55 & +3.90 \\
\midrule
REVE Large & 22 & 45.52 & 50.69 & +5.17 \\
REVE Base & 22 & 45.31 & 49.14 & +3.84 \\
LUNA Huge & 22 & 48.40 & 49.11 & +0.71 \\
LUNA Large & 22 & 46.34 & 47.59 & +1.25 \\
CSBrain & 22 & 40.89 & 48.68 & +7.79 \\
BIOT & 22 & 48.44 & 50.83 & +2.38 \\
LUNA Base & 22 & 43.79 & 46.72 & +2.92 \\
Official LaBraM Base & 22 & 43.65 & 51.19 & +7.55 \\
CodeBrain & 22 & 45.07 & 52.29 & +7.22 \\
FEMBA Base & 22 & 39.74 & 45.97 & +6.23 \\
NeuroLM & 22 & 47.92 & 50.46 & +2.54 \\
BrainOmni & 22 & 41.04 & 50.68 & +9.65 \\
NeuroGPT & 22 & 43.58 & 44.41 & +0.83 \\
EEGMamba & 22 & 40.28 & 46.68 & +6.40 \\
CBraMod & 22 & 39.91 & 50.72 & +10.81 \\
BENDR & 22 & 32.74 & 33.92 & +1.18 \\
\end{longtable}
\tabnote{SleepEDF is excluded from every paired comparison. All 26 encoders cover the same 22 tasks under both readouts and form the matched cohort used in the main analysis. Atom-H TUAB No-pool uses the reported trajectory mean.}
\endgroup

\subsection{FT--LP ranking agreement}
\label{app:ft-lp-ordering}

LP and Full-FT can favor different encoders. Among the 26 pretrained models, 80 of 325 pairs (24.6\%) reverse order when compared by their 23-task means. At the task level, the rate rises to 37.4\% (2,797 of 7,475 comparisons).

The extension encoders outperform the 16 baselines in both settings in 131 of 160 comparisons, while 27 favor the extension encoder in only one setting. Tables~\ref{tab:ft-lp-directional}--\ref{tab:ft-lp-by-task} show the overall, per-encoder, and per-task results.

\begingroup
\small
\setlength{\tabcolsep}{6pt}
\renewcommand{\arraystretch}{1.08}
\setlength{\LTpre}{6pt}
\setlength{\LTpost}{8pt}
\begin{longtable}{@{}lrr@{}}
\caption{FT and LP comparisons of extension encoders against baselines.}\label{tab:ft-lp-directional}\\
\toprule
Outcome & \shortstack{23-task mean\\Model pairs} & \shortstack{Individual tasks\\Pair--task comparisons} \\
\midrule
\endfirsthead
\caption[]{FT and LP comparisons of extension encoders against baselines. (continued).}\\
\toprule
Outcome & \shortstack{23-task mean\\Model pairs} & \shortstack{Individual tasks\\Pair--task comparisons} \\
\midrule
\endhead
\midrule
\multicolumn{3}{r@{}}{\emph{Continued on the next page.}}\\
\endfoot
\bottomrule
\endlastfoot
FT higher, LP higher & 131 & 1835 \\
FT higher, LP lower & 5 & 602 \\
FT lower, LP higher & 22 & 750 \\
FT lower, LP lower & 2 & 486 \\
Tie in either setting & 0 & 7 \\
\textbf{Total} & 160 & 3{,}680 \\
\end{longtable}
\endgroup

\begingroup
\small
\setlength{\tabcolsep}{6pt}
\renewcommand{\arraystretch}{1.08}
\setlength{\LTpre}{6pt}
\setlength{\LTpost}{8pt}
\begin{longtable}{@{}lrrrr@{}}
\caption{Ordering relative to the 16 baselines for each extension encoder. Scores average 23 tasks.}\label{tab:ft-lp-by-model}\\
\toprule
Encoder & \shortstack{FT higher\\LP higher} & \shortstack{FT higher\\LP lower} & \shortstack{FT lower\\LP higher} & \shortstack{FT lower\\LP lower} \\
\midrule
\endfirsthead
\caption[]{Ordering relative to the 16 baselines for each extension encoder. Scores average 23 tasks. (continued).}\\
\toprule
Encoder & \shortstack{FT higher\\LP higher} & \shortstack{FT higher\\LP lower} & \shortstack{FT lower\\LP higher} & \shortstack{FT lower\\LP lower} \\
\midrule
\endhead
\midrule
\multicolumn{5}{r@{}}{\emph{Continued on the next page.}}\\
\endfoot
\bottomrule
\endlastfoot
Quark-B & 7 & 0 & 8 & 1 \\
Quark-L & 9 & 0 & 7 & 0 \\
Quark-H & 11 & 1 & 4 & 0 \\
Quark-G & 16 & 0 & 0 & 0 \\
Atom-B & 12 & 1 & 3 & 0 \\
Atom-L & 16 & 0 & 0 & 0 \\
Atom-H & 16 & 0 & 0 & 0 \\
Molecule-B & 12 & 3 & 0 & 1 \\
Molecule-L & 16 & 0 & 0 & 0 \\
Molecule-H & 16 & 0 & 0 & 0 \\
\textbf{Total} & 131 & 5 & 22 & 2 \\
\end{longtable}
\endgroup

\begingroup
\small
\setlength{\tabcolsep}{6pt}
\renewcommand{\arraystretch}{1.08}
\setlength{\LTpre}{6pt}
\setlength{\LTpost}{8pt}
\begin{longtable}{@{}lrrrrr@{}}
\caption{Task-level comparisons of extension encoders against baselines. Each task contributes 160 comparisons.}\label{tab:ft-lp-by-task}\\
\toprule
Task & \shortstack{FT higher\\LP higher} & \shortstack{FT higher\\LP lower} & \shortstack{FT lower\\LP higher} & \shortstack{FT lower\\LP lower} & Ties \\
\midrule
\endfirsthead
\caption[]{Task-level comparisons of extension encoders against baselines. Each task contributes 160 comparisons. (continued).}\\
\toprule
Task & \shortstack{FT higher\\LP higher} & \shortstack{FT higher\\LP lower} & \shortstack{FT lower\\LP higher} & \shortstack{FT lower\\LP lower} & Ties \\
\midrule
\endhead
\midrule
\multicolumn{6}{r@{}}{\emph{Continued on the next page.}}\\
\endfoot
\bottomrule
\endlastfoot
AD65 & 107 & 33 & 10 & 10 & 0 \\
BCIC IV-1 & 33 & 36 & 63 & 24 & 4 \\
BCIC IV-2a & 60 & 7 & 78 & 13 & 2 \\
BETA SSVEP & 128 & 26 & 2 & 4 & 0 \\
EEGDenoiseNet & 82 & 16 & 13 & 49 & 0 \\
EEGMAT & 103 & 34 & 12 & 10 & 1 \\
FACED & 50 & 7 & 75 & 28 & 0 \\
HMC & 100 & 52 & 3 & 5 & 0 \\
ISRUC S2 & 72 & 5 & 73 & 10 & 0 \\
ISRUC S3 & 113 & 12 & 19 & 16 & 0 \\
LEMON extraversion & 74 & 65 & 10 & 11 & 0 \\
LEMON gender & 93 & 40 & 10 & 17 & 0 \\
Monitoring ErrP & 89 & 21 & 32 & 18 & 0 \\
PD31 & 45 & 7 & 84 & 24 & 0 \\
SEED & 82 & 15 & 28 & 35 & 0 \\
SEED-IV & 61 & 20 & 45 & 34 & 0 \\
SEED-V & 91 & 24 & 31 & 14 & 0 \\
SEED-VII & 57 & 20 & 34 & 49 & 0 \\
SHU MI & 44 & 27 & 68 & 21 & 0 \\
SleepEDF & 125 & 26 & 3 & 6 & 0 \\
TUAB & 94 & 51 & 3 & 12 & 0 \\
TUEV & 88 & 51 & 3 & 18 & 0 \\
TUSL & 44 & 7 & 51 & 58 & 0 \\
\textbf{Total} & 1835 & 602 & 750 & 486 & 7 \\
\end{longtable}
\endgroup

LP ranks encoders by the linear accessibility of their frozen features. Full-FT also depends on how those features adapt to the target task. The reversals show that a lead under LP can disappear after adaptation, so LP rankings alone are insufficient for choosing an encoder for fine-tuning.

Retaining patch structure improves mean LP accuracy by 4.87 points (Section~\ref{sec:linear-probing}). Table~\ref{tab:ft-lp-readout-ordering} examines whether this gain also brings encoder rankings closer to Full-FT. Across the same 26 encoders and 22 tasks, excluding SleepEDF, the model-level reversal rate rises from 21.8\% with Pooled to 30.2\% with No-pool. The task-level rate also rises, from 34.1\% to 38.5\%.

\begingroup
\small
\setlength{\tabcolsep}{6pt}
\renewcommand{\arraystretch}{1.08}
\setlength{\LTpre}{6pt}
\setlength{\LTpost}{8pt}
\begin{longtable}{@{}lrrrrr@{}}
\caption{FT--LP ordering reversals on the same 26 encoders and 22 tasks.}\label{tab:ft-lp-readout-ordering}\\
\toprule
LP readout & \shortstack{Mean-score\\Reversals} & Share & \shortstack{Taskwise\\Reversals} & Share & \shortstack{Taskwise\\ties} \\
\midrule
\endfirsthead
\caption[]{FT--LP ordering reversals on the same 26 encoders and 22 tasks. (continued).}\\
\toprule
LP readout & \shortstack{Mean-score\\Reversals} & Share & \shortstack{Taskwise\\Reversals} & Share & \shortstack{Taskwise\\ties} \\
\midrule
\endhead
\midrule
\multicolumn{6}{r@{}}{\emph{Continued on the next page.}}\\
\endfoot
\bottomrule
\endlastfoot
Pooled & 71 & 21.8\% & 2436 & 34.1\% & 45 \\
No-pool & 98 & 30.2\% & 2754 & 38.5\% & 15 \\
Per-task best readout & 83 & 25.5\% & 2699 & 37.7\% & 16 \\
\end{longtable}
\endgroup

Removing average pooling improves LP accuracy without resolving the ranking gap. LP still uses a frozen encoder and linear head, whereas Full-FT can change the representation for the target task. The persistence of the gap with No-pool reinforces the need to evaluate fine-tuning potential after adaptation, alongside frozen-feature performance.

\begingroup
\setlength{\parindent}{0pt}
\section{Results by Checkpoint and Dataset}
\label{app:checkpoint-results}
\begingroup\small Cells report three-seed mean balanced accuracy (\%). FT is full fine-tuning; LP uses the historical task-selected readout, with Pooled fixed for SleepEDF. Means cover 23 tasks; epochs are zero-indexed. A dash marks an unavailable result. Molecule-B uses a 50-epoch schedule.\par\endgroup
\par
\begingroup
\setlength{\tabcolsep}{3.3pt}
\renewcommand{\arraystretch}{1.0}
\captionsetup{font=small,skip=5pt,hypcap=false}
\begingroup
\fontsize{9}{10}\selectfont
\setlength{\LTpre}{6pt}
\setlength{\LTpost}{8pt}
\begin{longtable}{@{}lrrrrrrrr@{}}
\caption{Quark checkpoints. Balanced accuracy (\%).}\label{tab:checkpoint-quark}\\
\toprule
Dataset & \multicolumn{2}{c}{\shortstack{Quark-B\\ep49}} & \multicolumn{2}{c}{\shortstack{Quark-L\\ep49}} & \multicolumn{2}{c}{\shortstack{Quark-H\\ep49}} & \multicolumn{2}{c}{\shortstack{Quark-G\\ep29}} \\
 & FT & LP & FT & LP & FT & LP & FT & LP \\
\midrule
\endfirsthead
\caption[]{Quark checkpoints. Balanced accuracy (\%) (continued).}\\
\toprule
Dataset & \multicolumn{2}{c}{\shortstack{Quark-B\\ep49}} & \multicolumn{2}{c}{\shortstack{Quark-L\\ep49}} & \multicolumn{2}{c}{\shortstack{Quark-H\\ep49}} & \multicolumn{2}{c}{\shortstack{Quark-G\\ep29}} \\
 & FT & LP & FT & LP & FT & LP & FT & LP \\
\midrule
\endhead
\midrule
\multicolumn{9}{r@{}}{\emph{Continued on the next page.}}\\
\endfoot
\bottomrule
\endlastfoot
AD65 & 54.05 & 53.24 & 60.54 & 58.72 & 62.69 & 50.55 & 77.16 & 59.24 \\
PD31 & 32.35 & 58.37 & 34.36 & 56.46 & 43.74 & 49.47 & 43.46 & 51.20 \\
BCIC IV-1 & 52.33 & 50.17 & 49.50 & 52.17 & 51.00 & 51.17 & 53.50 & 50.50 \\
BCIC IV-2a & 34.43 & 45.37 & 39.06 & 43.46 & 40.39 & 38.83 & 41.67 & 42.36 \\
SHU MI & 60.83 & 58.58 & 55.23 & 58.61 & 52.95 & 58.33 & 54.53 & 61.43 \\
BETA SSVEP & 33.81 & 11.96 & 41.43 & 18.99 & 32.11 & 18.45 & 43.66 & 24.38 \\
Monitoring ErrP & 58.23 & 55.35 & 59.18 & 59.01 & 55.87 & 60.33 & 56.33 & 66.19 \\
EEGMAT & 72.69 & 62.66 & 63.59 & 64.82 & 68.06 & 65.59 & 72.99 & 60.80 \\
\end{longtable}

\endgroup
\par
\begingroup
\fontsize{9}{10}\selectfont
\setlength{\LTpre}{6pt}
\setlength{\LTpost}{8pt}
\begin{longtable}{@{}lrrrrrrrrrr@{}}
\caption{Atom-B: full-data checkpoints. Balanced accuracy (\%).}\label{tab:checkpoint-atom-base}\\
\toprule
Dataset & \multicolumn{2}{c}{\shortstack{Atom-B\\ep9}} & \multicolumn{2}{c}{\shortstack{Atom-B\\ep19}} & \multicolumn{2}{c}{\shortstack{Atom-B\\ep29}} & \multicolumn{2}{c}{\shortstack{Atom-B\\ep39}} & \multicolumn{2}{c}{\shortstack{Atom-B\\ep49}} \\
 & FT & LP & FT & LP & FT & LP & FT & LP & FT & LP \\
\midrule
\endfirsthead
\caption[]{Atom-B: full-data checkpoints. Balanced accuracy (\%) (continued).}\\
\toprule
Dataset & \multicolumn{2}{c}{\shortstack{Atom-B\\ep9}} & \multicolumn{2}{c}{\shortstack{Atom-B\\ep19}} & \multicolumn{2}{c}{\shortstack{Atom-B\\ep29}} & \multicolumn{2}{c}{\shortstack{Atom-B\\ep39}} & \multicolumn{2}{c}{\shortstack{Atom-B\\ep49}} \\
 & FT & LP & FT & LP & FT & LP & FT & LP & FT & LP \\
\midrule
\endhead
\midrule
\multicolumn{11}{r@{}}{\emph{Continued on the next page.}}\\
\endfoot
\bottomrule
\endlastfoot
AD65 & 57.90 & 53.42 & 56.77 & 55.98 & 67.88 & 53.72 & 59.53 & 53.58 & 55.52 & 52.63 \\
PD31 & 40.29 & 49.75 & 44.29 & 53.15 & 37.49 & 52.10 & 43.26 & 51.69 & 53.74 & 49.39 \\
BCIC IV-1 & 53.83 & 52.50 & 51.83 & 51.00 & 49.00 & 51.67 & 51.00 & 49.33 & 51.83 & 50.50 \\
BCIC IV-2a & 31.83 & 45.60 & 32.35 & 44.56 & 33.91 & 44.97 & 33.28 & 42.36 & 36.69 & 41.90 \\
SHU MI & 51.94 & 60.25 & 55.52 & 60.37 & 54.12 & 60.39 & 55.21 & 59.86 & 56.24 & 58.41 \\
BETA SSVEP & 37.08 & 5.45 & 38.78 & 6.43 & 34.94 & 10.24 & 35.77 & 13.18 & 31.76 & 11.64 \\
Monitoring ErrP & 55.69 & 53.42 & 58.31 & 54.33 & 56.32 & 58.87 & 55.53 & 57.62 & 56.05 & 57.69 \\
EEGMAT & 73.61 & 64.66 & 73.93 & 72.22 & 74.69 & 68.36 & 78.40 & 68.06 & 72.87 & 66.21 \\
\end{longtable}

\endgroup
\par
\begingroup
\fontsize{9}{10}\selectfont
\setlength{\LTpre}{6pt}
\setlength{\LTpost}{8pt}
\begin{longtable}{@{}lrrrrrrrrrr@{}}
\caption{Atom-L: full-data checkpoints. Balanced accuracy (\%).}\label{tab:checkpoint-atom-large}\\
\toprule
Dataset & \multicolumn{2}{c}{\shortstack{Atom-L\\ep9}} & \multicolumn{2}{c}{\shortstack{Atom-L\\ep19}} & \multicolumn{2}{c}{\shortstack{Atom-L\\ep29}} & \multicolumn{2}{c}{\shortstack{Atom-L\\ep39}} & \multicolumn{2}{c}{\shortstack{Atom-L\\ep49}} \\
 & FT & LP & FT & LP & FT & LP & FT & LP & FT & LP \\
\midrule
\endfirsthead
\caption[]{Atom-L: full-data checkpoints. Balanced accuracy (\%) (continued).}\\
\toprule
Dataset & \multicolumn{2}{c}{\shortstack{Atom-L\\ep9}} & \multicolumn{2}{c}{\shortstack{Atom-L\\ep19}} & \multicolumn{2}{c}{\shortstack{Atom-L\\ep29}} & \multicolumn{2}{c}{\shortstack{Atom-L\\ep39}} & \multicolumn{2}{c}{\shortstack{Atom-L\\ep49}} \\
 & FT & LP & FT & LP & FT & LP & FT & LP & FT & LP \\
\midrule
\endhead
\midrule
\multicolumn{11}{r@{}}{\emph{Continued on the next page.}}\\
\endfoot
\bottomrule
\endlastfoot
AD65 & 67.53 & 60.69 & 59.44 & 60.85 & 63.05 & 58.06 & 60.01 & 58.84 & 59.86 & 57.16 \\
PD31 & 48.67 & 50.98 & 39.92 & 51.31 & 55.48 & 54.63 & 42.84 & 53.48 & 37.12 & 53.52 \\
BCIC IV-1 & 47.50 & 51.67 & 50.67 & 50.00 & 50.00 & 50.00 & 51.83 & 49.00 & 53.83 & 52.33 \\
BCIC IV-2a & 41.38 & 43.75 & 36.40 & 42.77 & 42.65 & 41.90 & 34.49 & 43.17 & 34.32 & 41.61 \\
SHU MI & 57.31 & 59.28 & 56.99 & 59.26 & 57.37 & 60.46 & 52.11 & 60.43 & 52.60 & 58.97 \\
BETA SSVEP & 45.09 & 15.39 & 44.02 & 14.85 & 45.57 & 14.64 & 45.51 & 13.51 & 45.48 & 13.54 \\
Monitoring ErrP & 63.67 & 61.47 & 65.05 & 60.29 & 67.09 & 61.17 & 69.02 & 61.08 & 57.01 & 60.53 \\
EEGMAT & 77.32 & 69.60 & 75.62 & 64.36 & 80.40 & 65.90 & 72.38 & 66.21 & 77.93 & 64.35 \\
\end{longtable}

\endgroup
\par
\begingroup
\fontsize{9}{10}\selectfont
\setlength{\LTpre}{6pt}
\setlength{\LTpost}{8pt}
\begin{longtable}{@{}lrrrrrrrr@{}}
\caption{Atom-H: full-data checkpoints. Balanced accuracy (\%).}\label{tab:checkpoint-atom-huge}\\
\toprule
Dataset & \multicolumn{2}{c}{\shortstack{Atom-H\\ep9}} & \multicolumn{2}{c}{\shortstack{Atom-H\\ep19}} & \multicolumn{2}{c}{\shortstack{Atom-H\\ep29}} & \multicolumn{2}{c}{\shortstack{Atom-H\\ep39}} \\
 & FT & LP & FT & LP & FT & LP & FT & LP \\
\midrule
\endfirsthead
\caption[]{Atom-H: full-data checkpoints. Balanced accuracy (\%) (continued).}\\
\toprule
Dataset & \multicolumn{2}{c}{\shortstack{Atom-H\\ep9}} & \multicolumn{2}{c}{\shortstack{Atom-H\\ep19}} & \multicolumn{2}{c}{\shortstack{Atom-H\\ep29}} & \multicolumn{2}{c}{\shortstack{Atom-H\\ep39}} \\
 & FT & LP & FT & LP & FT & LP & FT & LP \\
\midrule
\endhead
\midrule
\multicolumn{9}{r@{}}{\emph{Continued on the next page.}}\\
\endfoot
\bottomrule
\endlastfoot
AD65 & 62.40 & 58.47 & 56.45 & 59.78 & 63.38 & 55.25 & 63.08 & 55.58 \\
PD31 & 52.90 & 57.47 & 60.47 & 56.73 & 44.77 & 56.07 & 55.50 & 54.73 \\
BCIC IV-1 & 51.83 & 50.83 & 50.33 & 51.33 & 51.33 & 54.00 & 47.50 & 53.83 \\
BCIC IV-2a & 39.58 & 40.91 & 38.72 & 41.09 & 35.36 & 42.01 & 37.27 & 41.78 \\
SHU MI & 53.24 & 59.23 & 52.05 & 58.40 & 51.10 & 58.22 & 51.21 & 60.09 \\
BETA SSVEP & 45.68 & 17.89 & 44.64 & 18.24 & 45.65 & 17.59 & 46.49 & 17.41 \\
Monitoring ErrP & 59.11 & 61.35 & 63.53 & 60.31 & 64.20 & 61.89 & 63.75 & 60.69 \\
EEGMAT & 71.76 & 65.43 & 79.63 & 66.06 & 72.69 & 65.28 & 77.79 & 64.51 \\
\end{longtable}

\endgroup
\par
\begingroup
\fontsize{9}{10}\selectfont
\setlength{\LTpre}{6pt}
\setlength{\LTpost}{8pt}
\begin{longtable}{@{}lrrrrrrrrrr@{}}
\caption{Molecule-B: 50-epoch run. Balanced accuracy (\%).}\label{tab:checkpoint-molecule-base50}\\
\toprule
Dataset & \multicolumn{2}{c}{\shortstack{Molecule-B\\ep9}} & \multicolumn{2}{c}{\shortstack{Molecule-B\\ep19}} & \multicolumn{2}{c}{\shortstack{Molecule-B\\ep29}} & \multicolumn{2}{c}{\shortstack{Molecule-B\\ep39}} & \multicolumn{2}{c}{\shortstack{Molecule-B\\ep49}} \\
 & FT & LP & FT & LP & FT & LP & FT & LP & FT & LP \\
\midrule
\endfirsthead
\caption[]{Molecule-B: 50-epoch run. Balanced accuracy (\%) (continued).}\\
\toprule
Dataset & \multicolumn{2}{c}{\shortstack{Molecule-B\\ep9}} & \multicolumn{2}{c}{\shortstack{Molecule-B\\ep19}} & \multicolumn{2}{c}{\shortstack{Molecule-B\\ep29}} & \multicolumn{2}{c}{\shortstack{Molecule-B\\ep39}} & \multicolumn{2}{c}{\shortstack{Molecule-B\\ep49}} \\
 & FT & LP & FT & LP & FT & LP & FT & LP & FT & LP \\
\midrule
\endhead
\midrule
\multicolumn{11}{r@{}}{\emph{Continued on the next page.}}\\
\endfoot
\bottomrule
\endlastfoot
AD65 & 51.14 & 54.89 & 55.85 & 52.43 & 60.97 & 52.50 & 60.87 & 50.75 & 57.62 & 49.21 \\
PD31 & 46.18 & 54.89 & 57.81 & 50.06 & 56.68 & 51.30 & 52.17 & 51.49 & 52.55 & 50.14 \\
BCIC IV-1 & 51.50 & 52.67 & 49.33 & 51.00 & 49.67 & 51.67 & 55.00 & 54.00 & 49.50 & 53.17 \\
BCIC IV-2a & 34.72 & 42.77 & 35.76 & 40.05 & 33.91 & 41.61 & 34.95 & 40.91 & 37.67 & 40.16 \\
SHU MI & 57.13 & 61.86 & 55.98 & 61.94 & 57.04 & 61.15 & 55.06 & 61.06 & 54.16 & 59.60 \\
BETA SSVEP & 39.73 & 14.35 & 39.82 & 15.12 & 41.46 & 14.20 & 38.36 & 13.57 & 39.17 & 13.33 \\
Monitoring ErrP & 64.12 & 61.61 & 58.02 & 61.39 & 64.36 & 61.94 & 62.84 & 62.50 & 56.86 & 62.58 \\
EEGMAT & 77.16 & 68.06 & 73.15 & 66.36 & 79.78 & 65.90 & 77.47 & 60.80 & 78.24 & 61.88 \\
\end{longtable}

\endgroup
\par
\begingroup
\fontsize{9}{10}\selectfont
\setlength{\LTpre}{6pt}
\setlength{\LTpost}{8pt}
\begin{longtable}{@{}lrrrrrrrr@{}}
\caption{Molecule-L and Molecule-H checkpoints. Balanced accuracy (\%).}\label{tab:checkpoint-molecule-large-huge}\\
\toprule
Dataset & \multicolumn{2}{c}{\shortstack{Molecule-L\\ep9}} & \multicolumn{2}{c}{\shortstack{Molecule-L\\ep19}} & \multicolumn{2}{c}{\shortstack{Molecule-L\\ep29}} & \multicolumn{2}{c}{\shortstack{Molecule-H\\ep9}} \\
 & FT & LP & FT & LP & FT & LP & FT & LP \\
\midrule
\endfirsthead
\caption[]{Molecule-L and Molecule-H checkpoints. Balanced accuracy (\%) (continued).}\\
\toprule
Dataset & \multicolumn{2}{c}{\shortstack{Molecule-L\\ep9}} & \multicolumn{2}{c}{\shortstack{Molecule-L\\ep19}} & \multicolumn{2}{c}{\shortstack{Molecule-L\\ep29}} & \multicolumn{2}{c}{\shortstack{Molecule-H\\ep9}} \\
 & FT & LP & FT & LP & FT & LP & FT & LP \\
\midrule
\endhead
\midrule
\multicolumn{9}{r@{}}{\emph{Continued on the next page.}}\\
\endfoot
\bottomrule
\endlastfoot
AD65 & 62.23 & 53.48 & 64.21 & 52.20 & 59.90 & 49.85 & 61.24 & 55.45 \\
PD31 & 45.50 & 55.65 & 54.04 & 53.54 & 55.55 & 54.70 & 58.12 & 53.88 \\
BCIC IV-1 & 51.67 & 50.67 & 50.50 & 54.00 & 51.33 & 53.50 & 48.83 & 53.67 \\
BCIC IV-2a & 40.91 & 42.94 & 46.82 & 42.59 & 34.14 & 42.59 & 35.88 & 43.06 \\
SHU MI & 60.11 & 59.08 & 54.30 & 60.93 & 54.83 & 60.91 & 59.28 & 60.72 \\
BETA SSVEP & 45.68 & 16.67 & 45.36 & 15.54 & 43.72 & 17.23 & 49.91 & 19.88 \\
Monitoring ErrP & 72.67 & 60.05 & 63.39 & 61.57 & 58.03 & 61.77 & 63.14 & 63.03 \\
EEGMAT & 74.07 & 68.21 & 77.62 & 68.98 & 75.62 & 65.74 & 74.54 & 67.29 \\
\end{longtable}

\endgroup
\par
\begingroup
\fontsize{9}{10}\selectfont
\setlength{\LTpre}{6pt}
\setlength{\LTpost}{8pt}
\begin{longtable}{@{}lrrrrrrrr@{}}
\caption{Atom-B: pretraining data fractions. Balanced accuracy (\%).}\label{tab:checkpoint-fractions-49k-base}\\
\toprule
Dataset & \multicolumn{2}{c}{\shortstack{10\% data\\ep49}} & \multicolumn{2}{c}{\shortstack{20\% data\\ep49}} & \multicolumn{2}{c}{\shortstack{50\% data\\ep49}} & \multicolumn{2}{c}{\shortstack{100\% data\\ep49}} \\
 & FT & LP & FT & LP & FT & LP & FT & LP \\
\midrule
\endfirsthead
\caption[]{Atom-B: pretraining data fractions. Balanced accuracy (\%) (continued).}\\
\toprule
Dataset & \multicolumn{2}{c}{\shortstack{10\% data\\ep49}} & \multicolumn{2}{c}{\shortstack{20\% data\\ep49}} & \multicolumn{2}{c}{\shortstack{50\% data\\ep49}} & \multicolumn{2}{c}{\shortstack{100\% data\\ep49}} \\
 & FT & LP & FT & LP & FT & LP & FT & LP \\
\midrule
\endhead
\midrule
\multicolumn{9}{r@{}}{\emph{Continued on the next page.}}\\
\endfoot
\bottomrule
\endlastfoot
AD65 & 61.07 & 53.73 & 54.39 & 54.88 & 65.82 & 59.06 & 55.52 & 52.63 \\
PD31 & 45.45 & 47.43 & 44.35 & 47.43 & 44.83 & 51.32 & 53.74 & 49.39 \\
BCIC IV-1 & 50.50 & 54.67 & 50.00 & 54.33 & 54.50 & 51.83 & 51.83 & 50.50 \\
BCIC IV-2a & 37.21 & 42.13 & 33.10 & 41.78 & 36.34 & 42.42 & 36.69 & 41.90 \\
SHU MI & 53.56 & 61.17 & 56.16 & 59.38 & 52.88 & 58.47 & 56.24 & 58.41 \\
BETA SSVEP & 33.42 & 10.36 & 29.70 & 12.20 & 37.65 & 14.52 & 31.76 & 11.64 \\
Monitoring ErrP & 61.61 & 59.76 & 62.28 & 58.74 & 56.74 & 60.80 & 56.05 & 57.69 \\
EEGMAT & 68.69 & 68.67 & 70.69 & 66.98 & 76.54 & 68.22 & 72.87 & 66.21 \\
\end{longtable}
\tabnote{The 100\% reference is repeated from the full-data checkpoint tables.}
\endgroup
\par
\begingroup
\fontsize{9}{10}\selectfont
\setlength{\LTpre}{6pt}
\setlength{\LTpost}{8pt}
\begin{longtable}{@{}lrrrrrrrr@{}}
\caption{Atom-L: pretraining data fractions. Balanced accuracy (\%).}\label{tab:checkpoint-fractions-49k-large}\\
\toprule
Dataset & \multicolumn{2}{c}{\shortstack{10\% data\\ep49}} & \multicolumn{2}{c}{\shortstack{20\% data\\ep49}} & \multicolumn{2}{c}{\shortstack{50\% data\\ep49}} & \multicolumn{2}{c}{\shortstack{100\% data\\ep49}} \\
 & FT & LP & FT & LP & FT & LP & FT & LP \\
\midrule
\endfirsthead
\caption[]{Atom-L: pretraining data fractions. Balanced accuracy (\%) (continued).}\\
\toprule
Dataset & \multicolumn{2}{c}{\shortstack{10\% data\\ep49}} & \multicolumn{2}{c}{\shortstack{20\% data\\ep49}} & \multicolumn{2}{c}{\shortstack{50\% data\\ep49}} & \multicolumn{2}{c}{\shortstack{100\% data\\ep49}} \\
 & FT & LP & FT & LP & FT & LP & FT & LP \\
\midrule
\endhead
\midrule
\multicolumn{9}{r@{}}{\emph{Continued on the next page.}}\\
\endfoot
\bottomrule
\endlastfoot
AD65 & 57.02 & 53.94 & 51.77 & 54.57 & 64.14 & 52.26 & 59.86 & 57.16 \\
PD31 & 45.61 & 49.27 & 45.10 & 51.89 & 50.22 & 50.95 & 37.12 & 53.52 \\
BCIC IV-1 & 48.83 & 53.00 & 50.17 & 52.33 & 52.00 & 53.00 & 53.83 & 52.33 \\
BCIC IV-2a & 40.97 & 43.06 & 33.39 & 41.61 & 35.53 & 41.49 & 34.32 & 41.61 \\
SHU MI & 53.34 & 58.10 & 51.20 & 58.94 & 52.67 & 59.44 & 52.60 & 58.97 \\
BETA SSVEP & 41.85 & 17.86 & 40.39 & 16.46 & 46.58 & 14.05 & 45.48 & 13.54 \\
Monitoring ErrP & 56.95 & 58.98 & 57.66 & 58.69 & 61.34 & 59.99 & 57.01 & 60.53 \\
EEGMAT & 66.83 & 64.66 & 69.45 & 69.61 & 68.99 & 63.74 & 77.93 & 64.35 \\
\end{longtable}
\tabnote{The 100\% reference is repeated from the full-data checkpoint tables.}
\endgroup
\par
\begingroup
\fontsize{9}{10}\selectfont
\setlength{\LTpre}{6pt}
\setlength{\LTpost}{8pt}
\begin{longtable}{@{}lrrrrrrrr@{}}
\caption{Molecule-B: pretraining data fractions. Balanced accuracy (\%).}\label{tab:checkpoint-fractions-79k-base}\\
\toprule
Dataset & \multicolumn{2}{c}{\shortstack{10\% data\\ep49}} & \multicolumn{2}{c}{\shortstack{20\% data\\ep49}} & \multicolumn{2}{c}{\shortstack{50\% data\\ep49}} & \multicolumn{2}{c}{\shortstack{100\% data\\ep49}} \\
 & FT & LP & FT & LP & FT & LP & FT & LP \\
\midrule
\endfirsthead
\caption[]{Molecule-B: pretraining data fractions. Balanced accuracy (\%) (continued).}\\
\toprule
Dataset & \multicolumn{2}{c}{\shortstack{10\% data\\ep49}} & \multicolumn{2}{c}{\shortstack{20\% data\\ep49}} & \multicolumn{2}{c}{\shortstack{50\% data\\ep49}} & \multicolumn{2}{c}{\shortstack{100\% data\\ep49}} \\
 & FT & LP & FT & LP & FT & LP & FT & LP \\
\midrule
\endhead
\midrule
\multicolumn{9}{r@{}}{\emph{Continued on the next page.}}\\
\endfoot
\bottomrule
\endlastfoot
AD65 & 53.32 & 57.48 & 54.38 & 54.06 & 60.13 & 53.60 & 57.62 & 49.21 \\
PD31 & 52.93 & 48.74 & 59.64 & 57.88 & 52.08 & 55.15 & 52.55 & 50.14 \\
BCIC IV-1 & 54.00 & 52.67 & 50.67 & 50.50 & 52.67 & 52.33 & 49.50 & 53.17 \\
BCIC IV-2a & 36.34 & 44.27 & 34.72 & 43.06 & 41.55 & 40.39 & 37.67 & 40.16 \\
SHU MI & 53.37 & 61.28 & 58.07 & 61.31 & 54.97 & 60.55 & 54.16 & 59.60 \\
BETA SSVEP & 38.04 & 10.27 & 40.27 & 12.29 & 36.82 & 7.20 & 39.17 & 13.33 \\
Monitoring ErrP & 63.82 & 60.78 & 57.08 & 62.34 & 57.82 & 59.13 & 56.86 & 62.58 \\
EEGMAT & 75.62 & 62.81 & 77.31 & 70.68 & 73.46 & 70.84 & 78.24 & 61.88 \\
\end{longtable}
\tabnote{The 100\% reference is repeated from the full-data checkpoint tables. Molecule-B uses the independent 50-epoch run.}
\endgroup
\par
\begingroup
\fontsize{9}{10}\selectfont
\setlength{\LTpre}{6pt}
\setlength{\LTpost}{8pt}
\begin{longtable}{@{}lrrrrrrrr@{}}
\caption{External encoders (1/4). Balanced accuracy (\%).~\cite{elouahidi2025reve,doner2025luna}}\label{tab:checkpoint-external-1}\\
\toprule
Dataset & \multicolumn{2}{c}{\shortstack{REVE\\Base}} & \multicolumn{2}{c}{\shortstack{REVE\\Large}} & \multicolumn{2}{c}{\shortstack{LUNA\\Base}} & \multicolumn{2}{c}{\shortstack{LUNA\\Large}} \\
 & FT & LP & FT & LP & FT & LP & FT & LP \\
\midrule
\endfirsthead
\caption[]{External encoders (1/4). Balanced accuracy (\%).~\cite{elouahidi2025reve,doner2025luna} (continued).}\\
\toprule
Dataset & \multicolumn{2}{c}{\shortstack{REVE\\Base}} & \multicolumn{2}{c}{\shortstack{REVE\\Large}} & \multicolumn{2}{c}{\shortstack{LUNA\\Base}} & \multicolumn{2}{c}{\shortstack{LUNA\\Large}} \\
 & FT & LP & FT & LP & FT & LP & FT & LP \\
\midrule
\endhead
\midrule
\multicolumn{9}{r@{}}{\emph{Continued on the next page.}}\\
\endfoot
\bottomrule
\endlastfoot
AD65 & 54.71 & 48.30 & 58.29 & 53.68 & 52.42 & 44.51 & 48.72 & 44.68 \\
PD31 & 40.54 & 49.78 & 50.56 & 52.25 & 40.76 & 40.43 & 53.66 & 41.63 \\
BCIC IV-1 & 54.17 & 52.00 & 53.67 & 53.17 & 52.00 & 51.17 & 51.67 & 52.67 \\
BCIC IV-2a & 41.84 & 38.02 & 44.33 & 44.73 & 38.25 & 35.82 & 37.21 & 36.57 \\
SHU MI & 62.35 & 59.11 & 56.50 & 58.93 & 58.93 & 58.27 & 62.70 & 55.80 \\
BETA SSVEP & 38.39 & 14.61 & 35.30 & 18.33 & 27.14 & 4.43 & 28.24 & 4.05 \\
Monitoring ErrP & 62.48 & 62.29 & 56.75 & 65.35 & 58.95 & 57.19 & 56.81 & 59.05 \\
EEGMAT & 69.92 & 60.04 & 74.85 & 64.67 & 63.16 & 59.47 & 56.92 & 58.83 \\
\end{longtable}

\endgroup
\par
\begingroup
\fontsize{9}{10}\selectfont
\setlength{\LTpre}{6pt}
\setlength{\LTpost}{8pt}
\begin{longtable}{@{}lrrrrrrrr@{}}
\caption{External encoders (2/4). Balanced accuracy (\%).~\cite{zhou2025csbrain,jiang2024labram,ma2026codebrain}}\label{tab:checkpoint-external-2}\\
\toprule
Dataset & \multicolumn{2}{c}{\shortstack{LUNA\\Huge}} & \multicolumn{2}{c}{CSBrain} & \multicolumn{2}{c}{\shortstack{Official LaBraM\\Base}} & \multicolumn{2}{c}{CodeBrain} \\
 & FT & LP & FT & LP & FT & LP & FT & LP \\
\midrule
\endfirsthead
\caption[]{External encoders (2/4). Balanced accuracy (\%).~\cite{zhou2025csbrain,jiang2024labram,ma2026codebrain} (continued).}\\
\toprule
Dataset & \multicolumn{2}{c}{\shortstack{LUNA\\Huge}} & \multicolumn{2}{c}{CSBrain} & \multicolumn{2}{c}{\shortstack{Official LaBraM\\Base}} & \multicolumn{2}{c}{CodeBrain} \\
 & FT & LP & FT & LP & FT & LP & FT & LP \\
\midrule
\endhead
\midrule
\multicolumn{9}{r@{}}{\emph{Continued on the next page.}}\\
\endfoot
\bottomrule
\endlastfoot
AD65 & 51.81 & 44.40 & 56.87 & 51.62 & 55.53 & 57.29 & 56.72 & 60.25 \\
PD31 & 54.95 & 48.42 & 53.58 & 41.26 & 55.64 & 46.23 & 51.58 & 49.95 \\
BCIC IV-1 & 48.67 & 51.83 & 53.17 & 52.00 & 50.67 & 49.33 & 49.17 & 49.50 \\
BCIC IV-2a & 41.38 & 35.42 & 36.17 & 38.37 & 37.27 & 39.70 & 37.38 & 42.19 \\
SHU MI & 61.12 & 58.42 & 56.67 & 55.59 & 51.64 & 58.46 & 50.94 & 60.57 \\
BETA SSVEP & 18.93 & 5.95 & 31.64 & 19.49 & 31.73 & 14.11 & 12.86 & 4.64 \\
Monitoring ErrP & 58.64 & 61.37 & 58.98 & 56.89 & 56.39 & 60.76 & 59.86 & 54.63 \\
EEGMAT & 66.36 & 60.81 & 64.06 & 56.51 & 73.92 & 67.29 & 67.14 & 65.59 \\
\end{longtable}

\endgroup
\par
\begingroup
\fontsize{9}{10}\selectfont
\setlength{\LTpre}{6pt}
\setlength{\LTpost}{8pt}
\begin{longtable}{@{}lrrrrrrrr@{}}
\caption{External encoders (3/4). Balanced accuracy (\%).~\cite{jiang2025neurolm,xiao2025brainomni,cui2023neurogpt,wang2025eegmamba}}\label{tab:checkpoint-external-3}\\
\toprule
Dataset & \multicolumn{2}{c}{NeuroLM} & \multicolumn{2}{c}{BrainOmni} & \multicolumn{2}{c}{NeuroGPT} & \multicolumn{2}{c}{EEGMamba} \\
 & FT & LP & FT & LP & FT & LP & FT & LP \\
\midrule
\endfirsthead
\caption[]{External encoders (3/4). Balanced accuracy (\%).~\cite{jiang2025neurolm,xiao2025brainomni,cui2023neurogpt,wang2025eegmamba} (continued).}\\
\toprule
Dataset & \multicolumn{2}{c}{NeuroLM} & \multicolumn{2}{c}{BrainOmni} & \multicolumn{2}{c}{NeuroGPT} & \multicolumn{2}{c}{EEGMamba} \\
 & FT & LP & FT & LP & FT & LP & FT & LP \\
\midrule
\endhead
\midrule
\multicolumn{9}{r@{}}{\emph{Continued on the next page.}}\\
\endfoot
\bottomrule
\endlastfoot
AD65 & 57.27 & 50.44 & 59.06 & 53.96 & 55.73 & 45.58 & 52.42 & 51.29 \\
PD31 & 58.74 & 61.57 & 60.07 & 48.76 & 56.44 & 34.44 & 49.21 & 50.92 \\
BCIC IV-1 & 51.17 & 50.67 & 49.33 & 54.33 & 52.00 & 53.67 & 51.50 & 49.83 \\
BCIC IV-2a & 35.65 & 40.74 & 45.43 & 38.95 & 36.11 & 39.18 & 32.81 & 37.21 \\
SHU MI & 57.13 & 58.75 & 57.43 & 58.61 & 54.59 & 51.90 & 50.55 & 58.70 \\
BETA SSVEP & 12.05 & 6.34 & 9.29 & 17.17 & 17.23 & 2.77 & 3.45 & 6.55 \\
Monitoring ErrP & 53.55 & 55.40 & 56.22 & 58.47 & 56.30 & 49.68 & 52.47 & 58.19 \\
EEGMAT & 68.21 & 66.67 & 50.00 & 66.68 & 62.81 & 49.69 & 50.31 & 55.41 \\
\end{longtable}

\endgroup
\par
\begingroup
\fontsize{9}{10}\selectfont
\setlength{\LTpre}{6pt}
\setlength{\LTpost}{8pt}
\begin{longtable}{@{}lrrrrrrrr@{}}
\caption{External encoders (4/4). Balanced accuracy (\%).~\cite{wang2025cbramod,kostas2021bendr,yang2023biot,tegon2025femba}}\label{tab:checkpoint-external-4}\\
\toprule
Dataset & \multicolumn{2}{c}{CBraMod} & \multicolumn{2}{c}{BENDR} & \multicolumn{2}{c}{BIOT} & \multicolumn{2}{c}{\shortstack{FEMBA\\Base}} \\
 & FT & LP & FT & LP & FT & LP & FT & LP \\
\midrule
\endfirsthead
\caption[]{External encoders (4/4). Balanced accuracy (\%).~\cite{wang2025cbramod,kostas2021bendr,yang2023biot,tegon2025femba} (continued).}\\
\toprule
Dataset & \multicolumn{2}{c}{CBraMod} & \multicolumn{2}{c}{BENDR} & \multicolumn{2}{c}{BIOT} & \multicolumn{2}{c}{\shortstack{FEMBA\\Base}} \\
 & FT & LP & FT & LP & FT & LP & FT & LP \\
\midrule
\endhead
\midrule
\multicolumn{9}{r@{}}{\emph{Continued on the next page.}}\\
\endfoot
\bottomrule
\endlastfoot
AD65 & 53.71 & 55.85 & 46.13 & 33.33 & 58.31 & 47.26 & 50.21 & 36.44 \\
PD31 & 39.91 & 44.44 & 66.67 & 33.33 & 69.07 & 61.92 & 54.02 & 47.86 \\
BCIC IV-1 & 55.33 & 49.50 & 51.17 & 52.00 & 49.83 & 52.33 & 50.50 & 51.00 \\
BCIC IV-2a & 30.15 & 40.16 & 29.86 & 27.14 & 40.45 & 33.04 & 41.09 & 37.09 \\
SHU MI & 50.81 & 61.85 & 52.81 & 50.46 & 53.91 & 53.53 & 58.98 & 60.02 \\
BETA SSVEP & 26.61 & 16.34 & 3.30 & 2.59 & 18.36 & 5.65 & 24.17 & 11.31 \\
Monitoring ErrP & 52.55 & 56.96 & 54.80 & 50.72 & 52.88 & 52.29 & 55.65 & 54.51 \\
EEGMAT & 62.09 & 62.35 & 64.52 & 50.00 & 66.67 & 62.22 & 70.70 & 50.00 \\
\end{longtable}
\tabnote{EEGDenoiseNet uses single-channel Cz input for BIOT and a Cz-only, zero-filled montage for FEMBA (Appendix~\ref{app:evaluation-settings}). BENDR FT uses the stable grid except on BCIC IV-2a.}
\endgroup
\par
\begingroup
\fontsize{9}{10}\selectfont
\setlength{\LTpre}{6pt}
\setlength{\LTpost}{8pt}
\begin{longtable}{@{}lrrrrrr@{}}
\caption{Models trained from random initialization. Balanced accuracy (\%).~\cite{lawhern2018eegnet,song2023eegconformer}}\label{tab:checkpoint-scratch}\\
\toprule
Dataset & \multicolumn{2}{c}{EEGNet} & \multicolumn{2}{c}{EEGConformer} & \multicolumn{2}{c}{\shortstack{LaBraM\\Base}} \\
 & FT & LP & FT & LP & FT & LP \\
\midrule
\endfirsthead
\caption[]{Models trained from random initialization. Balanced accuracy (\%).~\cite{lawhern2018eegnet,song2023eegconformer} (continued).}\\
\toprule
Dataset & \multicolumn{2}{c}{EEGNet} & \multicolumn{2}{c}{EEGConformer} & \multicolumn{2}{c}{\shortstack{LaBraM\\Base}} \\
 & FT & LP & FT & LP & FT & LP \\
\midrule
\endhead
\midrule
\multicolumn{7}{r@{}}{\emph{Continued on the next page.}}\\
\endfoot
\bottomrule
\endlastfoot
AD65 & 52.70 & \na & 58.60 & \na & 47.71 & \na \\
PD31 & 64.43 & \na & 52.05 & \na & 40.57 & \na \\
BCIC IV-1 & 51.83 & \na & 53.83 & \na & 54.17 & \na \\
BCIC IV-2a & 36.17 & \na & 50.29 & \na & 33.80 & \na \\
SHU MI & 57.81 & \na & 59.37 & \na & 51.62 & \na \\
BETA SSVEP & 5.60 & \na & 12.59 & \na & 5.09 & \na \\
Monitoring ErrP & 52.48 & \na & 63.42 & \na & 57.45 & \na \\
EEGMAT & 59.15 & \na & 71.91 & \na & 64.54 & \na \\
\end{longtable}
\tabnote{LP was not evaluated for these randomly initialized models.}
\endgroup
\par
\endgroup

\endgroup

\section{Pretraining Trajectories}
\label{app:training-dynamics}

The training records cover 22 runs and a separate two-epoch continuation of Molecule-L, totaling 941 recorded epochs. Figures~\ref{fig:training-quark}--\ref{fig:training-molecule} show shape-preserving cubic curves through the recorded losses, with interpolation in log-loss coordinates on logarithmic axes. Fractions refer to the retained window streams; all reduced-data runs have 50-epoch schedules. Some trajectories extend beyond the checkpoints evaluated downstream.

\par\medskip
\begingroup\centering
\captionsetup{type=figure,hypcap=false}
\includegraphics[width=0.93\linewidth]{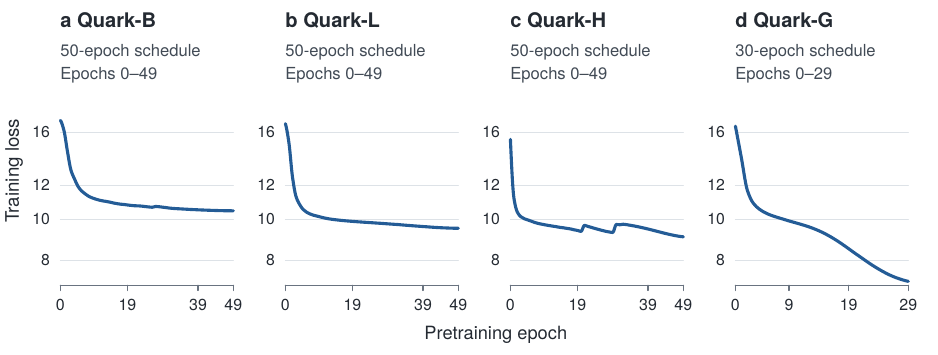}
\caption{Quark pretraining loss. Base, Large, and Huge use 50-epoch schedules; Giant uses 30 epochs. Each panel shows its full recorded range. Loss axes are logarithmic.}
\label{fig:training-quark}
\par\endgroup
\medskip

\subsection{Atom}

Atom's full-data and reduced-data runs use 50-epoch schedules. Figure~\ref{fig:training-atom} compares the full-data trajectories with the 10\%, 20\%, and 50\% data fractions at each encoder size.

\par\medskip
\begingroup\centering
\captionsetup{type=figure,hypcap=false}
\includegraphics[width=0.82\linewidth]{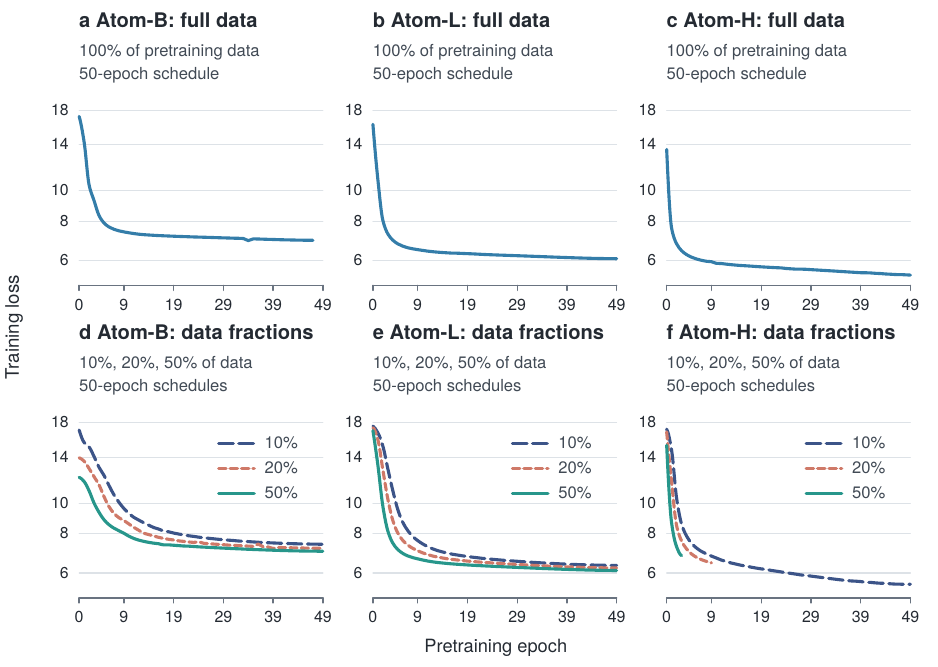}
\caption{Atom pretraining loss. Full-data runs appear separately from reduced-data runs. Loss axes are logarithmic.}
\label{fig:training-atom}
\par\endgroup
\medskip

\subsection{Molecule}

Molecule-B uses a 50-epoch schedule. The Large continuation resumes ep29 with a separate two-epoch rewarm schedule; it has no downstream evaluation. Huge has 20 recorded epochs under a 50-epoch schedule, with its last evaluated checkpoint at ep9. Tokenizers and input preparation differ across collections, limiting direct comparison of their absolute losses.

\par\medskip
\begingroup\centering
\captionsetup{type=figure,hypcap=false}
\includegraphics[width=0.82\linewidth]{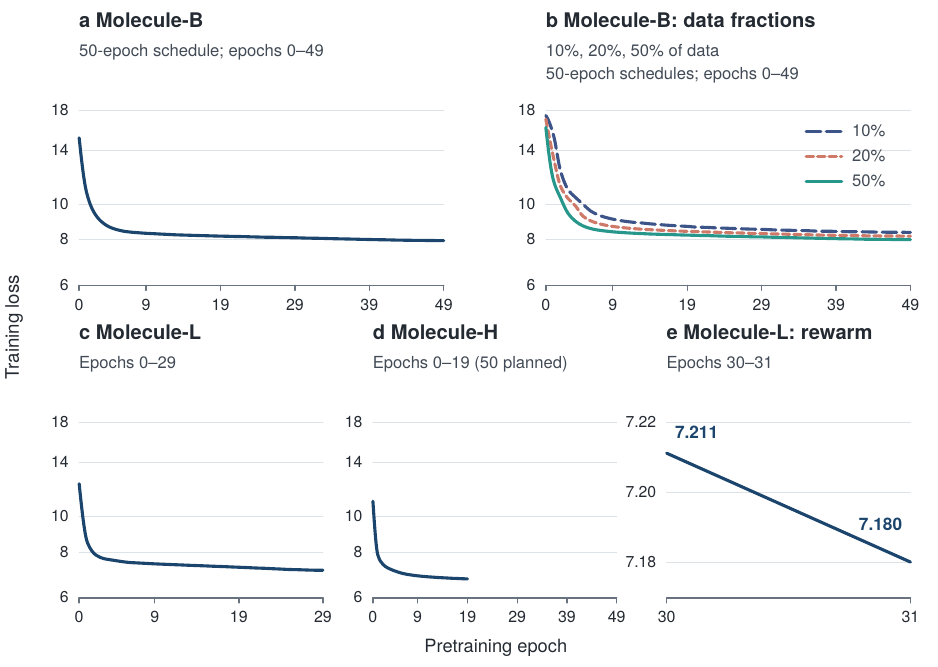}
\caption{Molecule pretraining loss for Base, reduced-data Base runs, Large, Huge, and the Large continuation. Loss axes are logarithmic except for the continuation.}
\label{fig:training-molecule}
\par\endgroup
\medskip

\section{Joint EEG--LLM Results}
\label{app:eeg-llm}
\begingroup\small
These experiments follow the joint-training protocol in Section~\ref{sec:eeg-llm}. Ten EEG encoders are evaluated with six Qwen3 sizes, giving 60 configurations and 180 runs. Tables report test balanced accuracy (\%) as mean $\pm$ sample SD over seeds 42, 10, and 5. The macro row averages the 23 tasks within each seed before aggregation. $\dagger$ISRUC S2 retains the original channel-cache defect and remains in the macro mean.
\par\endgroup
\begingroup
\captionsetup{skip=5pt}

\begingroup
\fontsize{9.5}{12}\selectfont
\setlength{\tabcolsep}{5pt}
\renewcommand{\arraystretch}{1.08}
\setlength{\LTpre}{6pt}
\setlength{\LTpost}{8pt}
\begin{longtable}{@{}>{\raggedright\arraybackslash}p{0.28\linewidth}p{0.68\linewidth}@{}}
\caption{Joint EEG--LLM training and evaluation settings.}\label{tab:eeg-llm-settings}\\
\toprule
Setting & Value \\
\midrule
\endfirsthead
\caption[]{Joint EEG--LLM training and evaluation settings (continued).}\\
\toprule
Setting & Value \\
\midrule
\endhead
\midrule
\multicolumn{2}{r@{}}{\emph{Continued on the next page.}}\\
\endfoot
\bottomrule
\endlastfoot
EEG encoder & Frozen; the checkpoint is identified in each result-table caption. \\
Input preparation & Quark/Atom: input scale $0.01$, no per-window z-scoring. Molecule: input scale $1$, per-window z-scoring. \\
Cached features & Temporal patch features averaged across channels, standardized using the corresponding task and seed's training-split statistics. \\
Temporal projector & 32 tokens; adaptive average pooling or linear interpolation for shorter sequences; two-layer GELU MLP. \\
Language model & Qwen3-0.6B, 1.7B, 4B, 8B, 14B, or 32B; original weights frozen. \\
Trainable modules & Projector and LoRA adapters, trained jointly from step 0. \\
LoRA & Rank $16$, $\alpha=32$, dropout $0.1$; q/k/v/o and gate/up/down projections. \\
Optimizer & AdamW; $\beta_1=0.9$, $\beta_2=0.98$. \\
\end{longtable}
\endgroup
\endgroup

\smallskip
The experiments share the data splits and 40-window subject--class training cap used for single-task evaluation (Appendix~\ref{app:evaluation-settings}). SleepEDF uses night identifiers as split units, allowing person-level overlap; EEGDenoiseNet uses a within-subject sample split. Validation and test use all available windows.
\begingroup
\captionsetup{skip=4pt}

\begingroup
\fontsize{9}{10.6}\selectfont
\setlength{\tabcolsep}{3pt}
\renewcommand{\arraystretch}{1.0}
\setlength{\LTpre}{6pt}
\setlength{\LTpost}{8pt}
\begin{longtable}{@{}lrrrrrr@{}}
\caption{EEG--LLM results for Quark-B (epoch 49).}\label{tab:eeg-llm-quark-b}\\
\toprule
Task & Qwen3-0.6B & Qwen3-1.7B & Qwen3-4B & Qwen3-8B & Qwen3-14B & Qwen3-32B \\
\midrule
\endfirsthead
\caption[]{EEG--LLM results for Quark-B (epoch 49) (continued).}\\
\toprule
Task & Qwen3-0.6B & Qwen3-1.7B & Qwen3-4B & Qwen3-8B & Qwen3-14B & Qwen3-32B \\
\midrule
\endhead
\midrule
\multicolumn{7}{r@{}}{\emph{Continued on the next page.}}\\
\endfoot
\bottomrule
\endlastfoot
AD65 & $49.05\,\pm\,5.89$ & $49.80\,\pm\,7.63$ & $51.20\,\pm\,12.17$ & $47.23\,\pm\,6.03$ & $48.56\,\pm\,5.33$ & $50.21\,\pm\,6.74$ \\
PD31 & $47.41\,\pm\,11.17$ & $39.68\,\pm\,15.09$ & $42.78\,\pm\,13.03$ & $43.71\,\pm\,14.85$ & $37.45\,\pm\,15.96$ & $43.03\,\pm\,15.88$ \\
BCIC IV-1 & $49.33\,\pm\,1.44$ & $48.83\,\pm\,4.04$ & $48.33\,\pm\,1.76$ & $48.00\,\pm\,3.28$ & $50.00\,\pm\,4.50$ & $51.50\,\pm\,2.29$ \\
BCIC IV-2a & $34.43\,\pm\,0.36$ & $34.49\,\pm\,0.82$ & $34.90\,\pm\,0.52$ & $35.53\,\pm\,0.96$ & $36.05\,\pm\,1.06$ & $35.24\,\pm\,2.39$ \\
SHU MI & $50.67\,\pm\,1.01$ & $51.27\,\pm\,1.28$ & $51.26\,\pm\,1.66$ & $52.43\,\pm\,0.70$ & $50.58\,\pm\,0.98$ & $50.92\,\pm\,1.06$ \\
BETA SSVEP & $2.41\,\pm\,0.00$ & $2.32\,\pm\,0.36$ & $2.50\,\pm\,0.47$ & $2.44\,\pm\,0.31$ & $2.35\,\pm\,0.44$ & $2.62\,\pm\,0.46$ \\
Monitoring ErrP & $52.45\,\pm\,3.31$ & $54.84\,\pm\,2.36$ & $52.08\,\pm\,2.95$ & $54.89\,\pm\,1.95$ & $56.07\,\pm\,3.49$ & $52.93\,\pm\,3.49$ \\
EEGMAT & $66.07\,\pm\,4.19$ & $65.46\,\pm\,6.58$ & $63.75\,\pm\,11.63$ & $63.90\,\pm\,5.70$ & $61.62\,\pm\,5.59$ & $66.54\,\pm\,9.28$ \\
\end{longtable}
\endgroup
\endgroup

\begingroup
\captionsetup{skip=4pt}

\begingroup
\fontsize{9}{10.6}\selectfont
\setlength{\tabcolsep}{3pt}
\renewcommand{\arraystretch}{1.0}
\setlength{\LTpre}{6pt}
\setlength{\LTpost}{8pt}
\begin{longtable}{@{}lrrrrrr@{}}
\caption{EEG--LLM results for Quark-L (epoch 49).}\label{tab:eeg-llm-quark-l}\\
\toprule
Task & Qwen3-0.6B & Qwen3-1.7B & Qwen3-4B & Qwen3-8B & Qwen3-14B & Qwen3-32B \\
\midrule
\endfirsthead
\caption[]{EEG--LLM results for Quark-L (epoch 49) (continued).}\\
\toprule
Task & Qwen3-0.6B & Qwen3-1.7B & Qwen3-4B & Qwen3-8B & Qwen3-14B & Qwen3-32B \\
\midrule
\endhead
\midrule
\multicolumn{7}{r@{}}{\emph{Continued on the next page.}}\\
\endfoot
\bottomrule
\endlastfoot
AD65 & $53.42\,\pm\,5.40$ & $53.84\,\pm\,7.42$ & $51.92\,\pm\,1.49$ & $49.66\,\pm\,2.50$ & $53.99\,\pm\,6.61$ & $55.75\,\pm\,9.04$ \\
PD31 & $52.54\,\pm\,15.45$ & $41.52\,\pm\,5.86$ & $43.11\,\pm\,7.80$ & $47.44\,\pm\,8.81$ & $46.80\,\pm\,9.87$ & $42.68\,\pm\,10.46$ \\
BCIC IV-1 & $50.67\,\pm\,1.26$ & $49.83\,\pm\,4.16$ & $48.50\,\pm\,3.28$ & $46.50\,\pm\,1.32$ & $48.33\,\pm\,1.26$ & $50.33\,\pm\,3.06$ \\
BCIC IV-2a & $35.24\,\pm\,3.22$ & $38.43\,\pm\,2.49$ & $36.11\,\pm\,4.21$ & $35.13\,\pm\,3.53$ & $31.08\,\pm\,2.52$ & $32.70\,\pm\,2.40$ \\
SHU MI & $50.91\,\pm\,1.81$ & $52.82\,\pm\,1.23$ & $51.61\,\pm\,0.99$ & $52.24\,\pm\,1.87$ & $49.59\,\pm\,0.74$ & $51.18\,\pm\,0.92$ \\
BETA SSVEP & $2.41\,\pm\,0.31$ & $2.80\,\pm\,0.22$ & $2.38\,\pm\,0.14$ & $2.95\,\pm\,0.82$ & $2.65\,\pm\,0.14$ & $2.47\,\pm\,0.19$ \\
Monitoring ErrP & $56.24\,\pm\,3.26$ & $55.35\,\pm\,2.06$ & $54.05\,\pm\,4.38$ & $55.04\,\pm\,4.44$ & $50.92\,\pm\,3.14$ & $55.43\,\pm\,5.20$ \\
EEGMAT & $65.59\,\pm\,7.47$ & $67.91\,\pm\,8.39$ & $69.46\,\pm\,3.01$ & $64.05\,\pm\,3.73$ & $64.98\,\pm\,6.37$ & $71.77\,\pm\,5.78$ \\
\end{longtable}
\endgroup
\endgroup

\begingroup
\captionsetup{skip=4pt}

\begingroup
\fontsize{9}{10.6}\selectfont
\setlength{\tabcolsep}{3pt}
\renewcommand{\arraystretch}{1.0}
\setlength{\LTpre}{6pt}
\setlength{\LTpost}{8pt}
\begin{longtable}{@{}lrrrrrr@{}}
\caption{EEG--LLM results for Quark-H (epoch 49).}\label{tab:eeg-llm-quark-h}\\
\toprule
Task & Qwen3-0.6B & Qwen3-1.7B & Qwen3-4B & Qwen3-8B & Qwen3-14B & Qwen3-32B \\
\midrule
\endfirsthead
\caption[]{EEG--LLM results for Quark-H (epoch 49) (continued).}\\
\toprule
Task & Qwen3-0.6B & Qwen3-1.7B & Qwen3-4B & Qwen3-8B & Qwen3-14B & Qwen3-32B \\
\midrule
\endhead
\midrule
\multicolumn{7}{r@{}}{\emph{Continued on the next page.}}\\
\endfoot
\bottomrule
\endlastfoot
AD65 & $51.60\,\pm\,10.18$ & $48.55\,\pm\,9.01$ & $45.63\,\pm\,6.49$ & $45.90\,\pm\,7.73$ & $45.91\,\pm\,1.00$ & $47.84\,\pm\,9.54$ \\
PD31 & $43.42\,\pm\,11.43$ & $49.70\,\pm\,9.15$ & $44.62\,\pm\,11.34$ & $50.82\,\pm\,11.61$ & $47.78\,\pm\,13.38$ & $44.00\,\pm\,10.67$ \\
BCIC IV-1 & $51.67\,\pm\,3.33$ & $48.83\,\pm\,3.88$ & $50.00\,\pm\,1.50$ & $52.50\,\pm\,1.32$ & $48.00\,\pm\,6.54$ & $49.33\,\pm\,0.58$ \\
BCIC IV-2a & $32.29\,\pm\,1.71$ & $34.03\,\pm\,3.93$ & $34.20\,\pm\,3.94$ & $33.62\,\pm\,1.97$ & $32.52\,\pm\,4.10$ & $30.21\,\pm\,1.59$ \\
SHU MI & $51.43\,\pm\,0.46$ & $52.50\,\pm\,1.37$ & $50.92\,\pm\,1.11$ & $53.28\,\pm\,0.76$ & $50.27\,\pm\,0.40$ & $51.47\,\pm\,0.70$ \\
BETA SSVEP & $2.35\,\pm\,0.34$ & $2.68\,\pm\,0.36$ & $2.71\,\pm\,0.36$ & $2.65\,\pm\,0.22$ & $2.86\,\pm\,0.24$ & $2.32\,\pm\,0.24$ \\
Monitoring ErrP & $54.85\,\pm\,1.59$ & $59.85\,\pm\,2.43$ & $57.43\,\pm\,3.39$ & $56.57\,\pm\,0.86$ & $59.05\,\pm\,3.51$ & $58.21\,\pm\,2.20$ \\
EEGMAT & $67.91\,\pm\,10.34$ & $63.28\,\pm\,10.27$ & $65.13\,\pm\,7.26$ & $68.69\,\pm\,10.51$ & $62.52\,\pm\,7.32$ & $61.59\,\pm\,10.48$ \\
\end{longtable}
\endgroup
\endgroup

\begingroup
\captionsetup{skip=4pt}

\begingroup
\fontsize{9}{10.6}\selectfont
\setlength{\tabcolsep}{3pt}
\renewcommand{\arraystretch}{1.0}
\setlength{\LTpre}{6pt}
\setlength{\LTpost}{8pt}
\begin{longtable}{@{}lrrrrrr@{}}
\caption{EEG--LLM results for Quark-G (epoch 29).}\label{tab:eeg-llm-quark-g}\\
\toprule
Task & Qwen3-0.6B & Qwen3-1.7B & Qwen3-4B & Qwen3-8B & Qwen3-14B & Qwen3-32B \\
\midrule
\endfirsthead
\caption[]{EEG--LLM results for Quark-G (epoch 29) (continued).}\\
\toprule
Task & Qwen3-0.6B & Qwen3-1.7B & Qwen3-4B & Qwen3-8B & Qwen3-14B & Qwen3-32B \\
\midrule
\endhead
\midrule
\multicolumn{7}{r@{}}{\emph{Continued on the next page.}}\\
\endfoot
\bottomrule
\endlastfoot
AD65 & $52.81\,\pm\,5.63$ & $51.54\,\pm\,5.15$ & $49.68\,\pm\,5.41$ & $50.71\,\pm\,5.96$ & $52.29\,\pm\,7.07$ & $49.15\,\pm\,6.69$ \\
PD31 & $41.25\,\pm\,7.70$ & $39.61\,\pm\,10.31$ & $38.41\,\pm\,11.48$ & $42.00\,\pm\,5.90$ & $41.28\,\pm\,5.71$ & $44.52\,\pm\,2.40$ \\
BCIC IV-1 & $52.50\,\pm\,1.80$ & $47.83\,\pm\,2.02$ & $45.33\,\pm\,5.51$ & $49.33\,\pm\,1.76$ & $50.67\,\pm\,1.61$ & $48.00\,\pm\,2.18$ \\
BCIC IV-2a & $31.25\,\pm\,1.08$ & $35.76\,\pm\,1.51$ & $31.83\,\pm\,2.53$ & $34.66\,\pm\,2.09$ & $36.28\,\pm\,0.92$ & $34.03\,\pm\,4.56$ \\
SHU MI & $54.74\,\pm\,1.03$ & $51.96\,\pm\,3.15$ & $51.94\,\pm\,1.34$ & $52.97\,\pm\,2.03$ & $51.40\,\pm\,2.22$ & $51.42\,\pm\,1.31$ \\
BETA SSVEP & $2.77\,\pm\,0.54$ & $3.07\,\pm\,0.65$ & $2.89\,\pm\,0.22$ & $3.36\,\pm\,1.00$ & $2.83\,\pm\,0.42$ & $2.74\,\pm\,0.49$ \\
Monitoring ErrP & $58.13\,\pm\,2.43$ & $55.15\,\pm\,4.37$ & $56.91\,\pm\,2.17$ & $56.84\,\pm\,4.17$ & $57.50\,\pm\,1.40$ & $54.07\,\pm\,3.37$ \\
EEGMAT & $63.75\,\pm\,10.88$ & $73.32\,\pm\,10.39$ & $70.43\,\pm\,8.40$ & $70.86\,\pm\,9.99$ & $65.62\,\pm\,14.90$ & $67.79\,\pm\,8.97$ \\
\end{longtable}
\endgroup
\endgroup

\begingroup
\captionsetup{skip=4pt}

\begingroup
\fontsize{9}{10.6}\selectfont
\setlength{\tabcolsep}{3pt}
\renewcommand{\arraystretch}{1.0}
\setlength{\LTpre}{6pt}
\setlength{\LTpost}{8pt}
\begin{longtable}{@{}lrrrrrr@{}}
\caption{EEG--LLM results for Atom-B (epoch 49).}\label{tab:eeg-llm-atom-b}\\
\toprule
Task & Qwen3-0.6B & Qwen3-1.7B & Qwen3-4B & Qwen3-8B & Qwen3-14B & Qwen3-32B \\
\midrule
\endfirsthead
\caption[]{EEG--LLM results for Atom-B (epoch 49) (continued).}\\
\toprule
Task & Qwen3-0.6B & Qwen3-1.7B & Qwen3-4B & Qwen3-8B & Qwen3-14B & Qwen3-32B \\
\midrule
\endhead
\midrule
\multicolumn{7}{r@{}}{\emph{Continued on the next page.}}\\
\endfoot
\bottomrule
\endlastfoot
AD65 & $49.30\,\pm\,5.27$ & $50.27\,\pm\,5.41$ & $54.80\,\pm\,10.89$ & $54.85\,\pm\,4.42$ & $46.25\,\pm\,6.19$ & $45.74\,\pm\,5.75$ \\
PD31 & $40.56\,\pm\,7.36$ & $43.19\,\pm\,7.66$ & $46.30\,\pm\,12.81$ & $47.04\,\pm\,5.87$ & $41.35\,\pm\,6.01$ & $45.32\,\pm\,6.07$ \\
BCIC IV-1 & $51.00\,\pm\,4.36$ & $49.67\,\pm\,4.04$ & $51.67\,\pm\,7.29$ & $50.00\,\pm\,1.80$ & $51.83\,\pm\,3.88$ & $48.67\,\pm\,3.21$ \\
BCIC IV-2a & $33.62\,\pm\,5.49$ & $33.80\,\pm\,0.72$ & $34.84\,\pm\,2.37$ & $34.03\,\pm\,2.34$ & $32.64\,\pm\,3.35$ & $33.10\,\pm\,5.28$ \\
SHU MI & $51.21\,\pm\,0.38$ & $50.42\,\pm\,1.24$ & $51.36\,\pm\,1.31$ & $50.51\,\pm\,1.91$ & $51.54\,\pm\,2.27$ & $49.88\,\pm\,0.89$ \\
BETA SSVEP & $2.77\,\pm\,0.24$ & $2.65\,\pm\,0.14$ & $2.44\,\pm\,0.31$ & $2.53\,\pm\,0.05$ & $2.68\,\pm\,0.45$ & $2.98\,\pm\,0.19$ \\
Monitoring ErrP & $54.74\,\pm\,1.98$ & $55.87\,\pm\,2.71$ & $57.01\,\pm\,2.60$ & $54.58\,\pm\,1.70$ & $56.48\,\pm\,0.57$ & $56.89\,\pm\,1.55$ \\
EEGMAT & $75.19\,\pm\,6.67$ & $68.24\,\pm\,8.37$ & $69.30\,\pm\,7.00$ & $70.11\,\pm\,12.01$ & $67.14\,\pm\,8.65$ & $71.16\,\pm\,2.26$ \\
\end{longtable}
\endgroup
\endgroup

\begingroup
\captionsetup{skip=4pt}

\begingroup
\fontsize{9}{10.6}\selectfont
\setlength{\tabcolsep}{3pt}
\renewcommand{\arraystretch}{1.0}
\setlength{\LTpre}{6pt}
\setlength{\LTpost}{8pt}
\begin{longtable}{@{}lrrrrrr@{}}
\caption{EEG--LLM results for Atom-L (epoch 49).}\label{tab:eeg-llm-atom-l-ep49}\\
\toprule
Task & Qwen3-0.6B & Qwen3-1.7B & Qwen3-4B & Qwen3-8B & Qwen3-14B & Qwen3-32B \\
\midrule
\endfirsthead
\caption[]{EEG--LLM results for Atom-L (epoch 49) (continued).}\\
\toprule
Task & Qwen3-0.6B & Qwen3-1.7B & Qwen3-4B & Qwen3-8B & Qwen3-14B & Qwen3-32B \\
\midrule
\endhead
\midrule
\multicolumn{7}{r@{}}{\emph{Continued on the next page.}}\\
\endfoot
\bottomrule
\endlastfoot
AD65 & $51.49\,\pm\,10.19$ & $52.42\,\pm\,6.16$ & $50.67\,\pm\,2.97$ & $49.93\,\pm\,4.99$ & $50.04\,\pm\,4.36$ & $48.40\,\pm\,1.47$ \\
PD31 & $39.16\,\pm\,3.23$ & $40.19\,\pm\,12.95$ & $42.85\,\pm\,9.57$ & $36.44\,\pm\,5.79$ & $41.83\,\pm\,8.94$ & $41.72\,\pm\,11.59$ \\
BCIC IV-1 & $50.00\,\pm\,1.32$ & $49.50\,\pm\,2.18$ & $48.17\,\pm\,3.01$ & $46.83\,\pm\,1.04$ & $51.33\,\pm\,1.26$ & $50.67\,\pm\,0.76$ \\
BCIC IV-2a & $35.01\,\pm\,1.76$ & $35.13\,\pm\,2.37$ & $32.64\,\pm\,2.87$ & $33.91\,\pm\,4.02$ & $35.76\,\pm\,2.46$ & $33.28\,\pm\,1.86$ \\
SHU MI & $51.56\,\pm\,0.92$ & $51.89\,\pm\,0.62$ & $51.43\,\pm\,1.15$ & $51.96\,\pm\,1.47$ & $51.37\,\pm\,0.91$ & $51.08\,\pm\,0.44$ \\
BETA SSVEP & $2.86\,\pm\,0.78$ & $2.86\,\pm\,0.78$ & $2.47\,\pm\,0.05$ & $2.74\,\pm\,0.10$ & $2.83\,\pm\,0.49$ & $2.38\,\pm\,0.05$ \\
Monitoring ErrP & $54.32\,\pm\,2.21$ & $56.56\,\pm\,0.80$ & $53.19\,\pm\,2.09$ & $54.62\,\pm\,2.55$ & $54.68\,\pm\,2.93$ & $53.30\,\pm\,3.20$ \\
EEGMAT & $70.23\,\pm\,6.54$ & $72.55\,\pm\,3.22$ & $66.84\,\pm\,11.51$ & $64.68\,\pm\,3.62$ & $63.44\,\pm\,5.09$ & $69.02\,\pm\,3.73$ \\
\end{longtable}
\endgroup
\endgroup

\begingroup
\captionsetup{skip=4pt}

\begingroup
\fontsize{9}{10.6}\selectfont
\setlength{\tabcolsep}{3pt}
\renewcommand{\arraystretch}{1.0}
\setlength{\LTpre}{6pt}
\setlength{\LTpost}{8pt}
\begin{longtable}{@{}lrrrrrr@{}}
\caption{EEG--LLM results for Atom-H (epoch 49).}\label{tab:eeg-llm-atom-h}\\
\toprule
Task & Qwen3-0.6B & Qwen3-1.7B & Qwen3-4B & Qwen3-8B & Qwen3-14B & Qwen3-32B \\
\midrule
\endfirsthead
\caption[]{EEG--LLM results for Atom-H (epoch 49) (continued).}\\
\toprule
Task & Qwen3-0.6B & Qwen3-1.7B & Qwen3-4B & Qwen3-8B & Qwen3-14B & Qwen3-32B \\
\midrule
\endhead
\midrule
\multicolumn{7}{r@{}}{\emph{Continued on the next page.}}\\
\endfoot
\bottomrule
\endlastfoot
AD65 & $54.43\,\pm\,5.61$ & $53.29\,\pm\,6.21$ & $52.31\,\pm\,9.07$ & $57.39\,\pm\,9.42$ & $50.73\,\pm\,6.36$ & $53.02\,\pm\,9.06$ \\
PD31 & $45.29\,\pm\,10.84$ & $46.65\,\pm\,10.20$ & $44.51\,\pm\,4.30$ & $48.46\,\pm\,8.25$ & $49.44\,\pm\,5.88$ & $41.32\,\pm\,12.30$ \\
BCIC IV-1 & $50.83\,\pm\,0.58$ & $51.17\,\pm\,3.06$ & $48.50\,\pm\,1.80$ & $52.50\,\pm\,4.58$ & $49.50\,\pm\,3.91$ & $50.50\,\pm\,3.50$ \\
BCIC IV-2a & $32.47\,\pm\,3.38$ & $34.43\,\pm\,0.70$ & $35.59\,\pm\,2.30$ & $34.49\,\pm\,1.94$ & $34.78\,\pm\,4.20$ & $28.65\,\pm\,2.56$ \\
SHU MI & $51.07\,\pm\,0.59$ & $51.59\,\pm\,0.96$ & $50.82\,\pm\,1.68$ & $50.40\,\pm\,0.46$ & $52.21\,\pm\,0.84$ & $51.48\,\pm\,0.94$ \\
BETA SSVEP & $2.44\,\pm\,0.19$ & $3.69\,\pm\,0.95$ & $2.74\,\pm\,0.34$ & $2.86\,\pm\,0.18$ & $3.45\,\pm\,0.67$ & $3.01\,\pm\,0.14$ \\
Monitoring ErrP & $54.08\,\pm\,3.30$ & $57.07\,\pm\,3.52$ & $57.40\,\pm\,4.72$ & $56.56\,\pm\,1.38$ & $56.16\,\pm\,3.38$ & $58.08\,\pm\,3.53$ \\
EEGMAT & $70.68\,\pm\,9.75$ & $71.31\,\pm\,7.63$ & $70.24\,\pm\,5.76$ & $70.53\,\pm\,10.29$ & $69.46\,\pm\,10.86$ & $67.14\,\pm\,3.03$ \\
\end{longtable}
\endgroup
\endgroup

\begingroup
\captionsetup{skip=4pt}

\begingroup
\fontsize{9}{10.6}\selectfont
\setlength{\tabcolsep}{3pt}
\renewcommand{\arraystretch}{1.0}
\setlength{\LTpre}{6pt}
\setlength{\LTpost}{8pt}
\begin{longtable}{@{}lrrrrrr@{}}
\caption{EEG--LLM results for Molecule-B (epoch 49).}\label{tab:eeg-llm-molecule-b}\\
\toprule
Task & Qwen3-0.6B & Qwen3-1.7B & Qwen3-4B & Qwen3-8B & Qwen3-14B & Qwen3-32B \\
\midrule
\endfirsthead
\caption[]{EEG--LLM results for Molecule-B (epoch 49) (continued).}\\
\toprule
Task & Qwen3-0.6B & Qwen3-1.7B & Qwen3-4B & Qwen3-8B & Qwen3-14B & Qwen3-32B \\
\midrule
\endhead
\midrule
\multicolumn{7}{r@{}}{\emph{Continued on the next page.}}\\
\endfoot
\bottomrule
\endlastfoot
AD65 & $46.89\,\pm\,4.30$ & $45.28\,\pm\,4.39$ & $43.48\,\pm\,0.90$ & $45.92\,\pm\,1.18$ & $42.74\,\pm\,3.92$ & $46.25\,\pm\,2.28$ \\
PD31 & $48.05\,\pm\,8.00$ & $44.80\,\pm\,7.41$ & $43.44\,\pm\,7.11$ & $44.52\,\pm\,5.00$ & $44.61\,\pm\,6.43$ & $46.16\,\pm\,7.66$ \\
BCIC IV-1 & $51.33\,\pm\,0.58$ & $46.83\,\pm\,3.40$ & $50.67\,\pm\,1.89$ & $48.67\,\pm\,2.36$ & $48.83\,\pm\,2.36$ & $46.33\,\pm\,6.79$ \\
BCIC IV-2a & $36.52\,\pm\,4.34$ & $33.10\,\pm\,1.26$ & $33.91\,\pm\,2.70$ & $34.03\,\pm\,3.81$ & $33.51\,\pm\,3.30$ & $32.23\,\pm\,0.72$ \\
SHU MI & $51.70\,\pm\,1.29$ & $53.61\,\pm\,0.97$ & $52.05\,\pm\,1.73$ & $51.47\,\pm\,0.89$ & $50.09\,\pm\,0.26$ & $50.78\,\pm\,1.21$ \\
BETA SSVEP & $2.41\,\pm\,0.56$ & $2.53\,\pm\,0.34$ & $2.89\,\pm\,0.10$ & $2.74\,\pm\,0.14$ & $2.92\,\pm\,0.22$ & $2.50\,\pm\,0.63$ \\
Monitoring ErrP & $56.57\,\pm\,2.39$ & $57.89\,\pm\,2.33$ & $54.45\,\pm\,2.46$ & $54.21\,\pm\,4.60$ & $56.13\,\pm\,1.98$ & $53.53\,\pm\,1.76$ \\
EEGMAT & $65.60\,\pm\,9.36$ & $62.21\,\pm\,8.82$ & $60.81\,\pm\,8.71$ & $63.30\,\pm\,9.95$ & $64.52\,\pm\,12.21$ & $67.46\,\pm\,10.67$ \\
\end{longtable}
\endgroup
\endgroup

\begingroup
\captionsetup{skip=4pt}

\begingroup
\fontsize{9}{10.6}\selectfont
\setlength{\tabcolsep}{3pt}
\renewcommand{\arraystretch}{1.0}
\setlength{\LTpre}{6pt}
\setlength{\LTpost}{8pt}
\begin{longtable}{@{}lrrrrrr@{}}
\caption{EEG--LLM results for Molecule-L (epoch 29).}\label{tab:eeg-llm-molecule-l}\\
\toprule
Task & Qwen3-0.6B & Qwen3-1.7B & Qwen3-4B & Qwen3-8B & Qwen3-14B & Qwen3-32B \\
\midrule
\endfirsthead
\caption[]{EEG--LLM results for Molecule-L (epoch 29) (continued).}\\
\toprule
Task & Qwen3-0.6B & Qwen3-1.7B & Qwen3-4B & Qwen3-8B & Qwen3-14B & Qwen3-32B \\
\midrule
\endhead
\midrule
\multicolumn{7}{r@{}}{\emph{Continued on the next page.}}\\
\endfoot
\bottomrule
\endlastfoot
AD65 & $47.28\,\pm\,4.37$ & $45.84\,\pm\,10.38$ & $47.20\,\pm\,6.55$ & $48.55\,\pm\,3.23$ & $47.89\,\pm\,1.52$ & $51.18\,\pm\,3.29$ \\
PD31 & $47.18\,\pm\,7.62$ & $52.61\,\pm\,5.62$ & $48.94\,\pm\,5.80$ & $48.48\,\pm\,4.93$ & $47.91\,\pm\,9.01$ & $49.00\,\pm\,7.47$ \\
BCIC IV-1 & $47.50\,\pm\,1.80$ & $50.00\,\pm\,2.65$ & $49.33\,\pm\,1.89$ & $45.83\,\pm\,5.13$ & $50.67\,\pm\,1.26$ & $49.67\,\pm\,1.89$ \\
BCIC IV-2a & $35.47\,\pm\,4.37$ & $36.17\,\pm\,2.54$ & $38.60\,\pm\,1.58$ & $36.00\,\pm\,1.58$ & $34.49\,\pm\,2.34$ & $33.51\,\pm\,1.67$ \\
SHU MI & $52.05\,\pm\,3.46$ & $52.73\,\pm\,0.59$ & $52.10\,\pm\,2.00$ & $52.90\,\pm\,2.36$ & $52.39\,\pm\,1.63$ & $51.39\,\pm\,1.08$ \\
BETA SSVEP & $3.27\,\pm\,0.41$ & $2.83\,\pm\,0.27$ & $2.77\,\pm\,0.67$ & $3.12\,\pm\,0.18$ & $3.24\,\pm\,0.44$ & $2.56\,\pm\,0.10$ \\
Monitoring ErrP & $55.17\,\pm\,2.94$ & $54.99\,\pm\,3.10$ & $56.75\,\pm\,1.94$ & $53.46\,\pm\,3.71$ & $56.63\,\pm\,1.21$ & $58.47\,\pm\,3.99$ \\
EEGMAT & $64.67\,\pm\,14.35$ & $68.06\,\pm\,5.91$ & $71.31\,\pm\,3.95$ & $64.52\,\pm\,12.43$ & $62.05\,\pm\,4.55$ & $68.23\,\pm\,8.38$ \\
\end{longtable}
\endgroup
\endgroup

\begingroup
\captionsetup{skip=4pt}

\begingroup
\fontsize{9}{10.6}\selectfont
\setlength{\tabcolsep}{3pt}
\renewcommand{\arraystretch}{1.0}
\setlength{\LTpre}{6pt}
\setlength{\LTpost}{8pt}
\begin{longtable}{@{}lrrrrrr@{}}
\caption{EEG--LLM results for Molecule-H (epoch 9).}\label{tab:eeg-llm-molecule-h}\\
\toprule
Task & Qwen3-0.6B & Qwen3-1.7B & Qwen3-4B & Qwen3-8B & Qwen3-14B & Qwen3-32B \\
\midrule
\endfirsthead
\caption[]{EEG--LLM results for Molecule-H (epoch 9) (continued).}\\
\toprule
Task & Qwen3-0.6B & Qwen3-1.7B & Qwen3-4B & Qwen3-8B & Qwen3-14B & Qwen3-32B \\
\midrule
\endhead
\midrule
\multicolumn{7}{r@{}}{\emph{Continued on the next page.}}\\
\endfoot
\bottomrule
\endlastfoot
AD65 & $53.50\,\pm\,4.61$ & $48.32\,\pm\,5.74$ & $49.69\,\pm\,1.72$ & $50.44\,\pm\,8.14$ & $49.30\,\pm\,0.49$ & $50.70\,\pm\,3.55$ \\
PD31 & $43.78\,\pm\,6.05$ & $45.64\,\pm\,2.59$ & $45.79\,\pm\,8.54$ & $45.41\,\pm\,9.53$ & $44.06\,\pm\,12.60$ & $47.94\,\pm\,11.01$ \\
BCIC IV-1 & $52.00\,\pm\,2.78$ & $46.83\,\pm\,2.25$ & $52.83\,\pm\,5.39$ & $49.33\,\pm\,3.51$ & $48.50\,\pm\,1.32$ & $51.83\,\pm\,2.52$ \\
BCIC IV-2a & $35.94\,\pm\,2.18$ & $36.34\,\pm\,0.20$ & $34.78\,\pm\,6.36$ & $38.19\,\pm\,4.96$ & $34.95\,\pm\,5.78$ & $35.71\,\pm\,1.82$ \\
SHU MI & $53.79\,\pm\,2.60$ & $51.47\,\pm\,2.28$ & $51.01\,\pm\,0.69$ & $51.30\,\pm\,2.53$ & $49.58\,\pm\,1.27$ & $51.27\,\pm\,2.98$ \\
BETA SSVEP & $3.84\,\pm\,1.03$ & $3.69\,\pm\,0.74$ & $3.54\,\pm\,0.74$ & $3.24\,\pm\,0.27$ & $3.12\,\pm\,0.59$ & $3.42\,\pm\,0.21$ \\
Monitoring ErrP & $54.66\,\pm\,3.43$ & $57.80\,\pm\,3.03$ & $57.12\,\pm\,3.45$ & $56.25\,\pm\,3.91$ & $54.18\,\pm\,2.21$ & $58.06\,\pm\,1.28$ \\
EEGMAT & $70.84\,\pm\,5.20$ & $69.76\,\pm\,6.16$ & $69.45\,\pm\,9.85$ & $71.77\,\pm\,10.23$ & $65.29\,\pm\,14.12$ & $65.44\,\pm\,4.19$ \\
\end{longtable}
\endgroup
\endgroup

\subsection{Training-set size}
Table~\ref{tab:eeg-llm-train-samples} gives the training-window counts shared by the EEG--LLM configurations. Sampling is applied to training only, with at most 40 windows per registered subject--class unit. The split conventions are described in Appendix~\ref{app:evaluation-settings}; Section~\ref{sec:eeg-llm} gives the joint-training settings.
\begingroup
\captionsetup{skip=4pt}

\begingroup
\fontsize{9}{12.0}\selectfont
\setlength{\tabcolsep}{3pt}
\renewcommand{\arraystretch}{1.0}
\setlength{\LTpre}{6pt}
\setlength{\LTpost}{8pt}
\begin{longtable}{@{}lrrrrrr@{}}
\caption{Training windows before and after the per-subject/class cap.}\label{tab:eeg-llm-train-samples}\\
\toprule
& \multicolumn{2}{c}{Seed42} & \multicolumn{2}{c}{Seed10} & \multicolumn{2}{c}{Seed5} \\
\cmidrule(lr){2-3}\cmidrule(lr){4-5}\cmidrule(l){6-7}
Task & Before $\rightarrow$ after & Retained & Before $\rightarrow$ after & Retained & Before $\rightarrow$ after & Retained \\
\midrule
\endfirsthead
\caption[]{Training windows before and after the per-subject/class cap (continued).}\\
\toprule
& \multicolumn{2}{c}{Seed42} & \multicolumn{2}{c}{Seed10} & \multicolumn{2}{c}{Seed5} \\
\cmidrule(lr){2-3}\cmidrule(lr){4-5}\cmidrule(l){6-7}
Task & Before $\rightarrow$ after & Retained & Before $\rightarrow$ after & Retained & Before $\rightarrow$ after & Retained \\
\midrule
\endhead
\midrule
\multicolumn{7}{r@{}}{\emph{Continued on the next page.}}\\
\endfoot
\bottomrule
\endlastfoot
AD65 & 5,500\,$\rightarrow$\,2,790 & 50.7\% & 5,572\,$\rightarrow$\,2,790 & 50.1\% & 5,609\,$\rightarrow$\,2,790 & 49.7\% \\
PD31 & 7,710\,$\rightarrow$\,1,000 & 13.0\% & 7,354\,$\rightarrow$\,1,000 & 13.6\% & 7,544\,$\rightarrow$\,1,000 & 13.3\% \\
BCIC IV-1 & 1,000\,$\rightarrow$\,400 & 40.0\% & 1,000\,$\rightarrow$\,400 & 40.0\% & 1,000\,$\rightarrow$\,400 & 40.0\% \\
BCIC IV-2a & 4,032\,$\rightarrow$\,1,120 & 27.8\% & 4,032\,$\rightarrow$\,1,120 & 27.8\% & 4,032\,$\rightarrow$\,1,120 & 27.8\% \\
SHU MI & 9,556\,$\rightarrow$\,1,600 & 16.7\% & 9,559\,$\rightarrow$\,1,600 & 16.7\% & 9,575\,$\rightarrow$\,1,600 & 16.7\% \\
BETA SSVEP & 8,960\,$\rightarrow$\,8,960 & 100.0\% & 8,960\,$\rightarrow$\,8,960 & 100.0\% & 8,960\,$\rightarrow$\,8,960 & 100.0\% \\
Monitoring ErrP & 4,313\,$\rightarrow$\,320 & 7.4\% & 4,291\,$\rightarrow$\,320 & 7.5\% & 4,291\,$\rightarrow$\,320 & 7.5\% \\
EEGMAT & 1,371\,$\rightarrow$\,1,371 & 100.0\% & 1,380\,$\rightarrow$\,1,380 & 100.0\% & 1,372\,$\rightarrow$\,1,372 & 100.0\% \\
LEMON gender & 152,792\,$\rightarrow$\,6,480 & 4.2\% & 152,706\,$\rightarrow$\,6,480 & 4.2\% & 152,618\,$\rightarrow$\,6,480 & 4.2\% \\
LEMON extraversion & 152,792\,$\rightarrow$\,6,480 & 4.2\% & 152,706\,$\rightarrow$\,6,480 & 4.2\% & 152,618\,$\rightarrow$\,6,480 & 4.2\% \\
FACED & 8,232\,$\rightarrow$\,8,232 & 100.0\% & 8,232\,$\rightarrow$\,8,232 & 100.0\% & 8,232\,$\rightarrow$\,8,232 & 100.0\% \\
SEED & 11,916\,$\rightarrow$\,1,440 & 12.1\% & 11,916\,$\rightarrow$\,1,440 & 12.1\% & 11,916\,$\rightarrow$\,1,440 & 12.1\% \\
SEED-IV & 30,060\,$\rightarrow$\,1,920 & 6.4\% & 30,060\,$\rightarrow$\,1,920 & 6.4\% & 30,060\,$\rightarrow$\,1,920 & 6.4\% \\
SEED-V & 95,667\,$\rightarrow$\,2,600 & 2.7\% & 95,667\,$\rightarrow$\,2,600 & 2.7\% & 95,667\,$\rightarrow$\,2,600 & 2.7\% \\
SEED-VII & 21,872\,$\rightarrow$\,4,480 & 20.5\% & 21,872\,$\rightarrow$\,4,480 & 20.5\% & 21,872\,$\rightarrow$\,4,480 & 20.5\% \\
ISRUC S2 & 10,446\,$\rightarrow$\,1,200 & 11.5\% & 10,446\,$\rightarrow$\,1,200 & 11.5\% & 10,581\,$\rightarrow$\,1,200 & 11.3\% \\
ISRUC S3 & 7,058\,$\rightarrow$\,1,600 & 22.7\% & 6,901\,$\rightarrow$\,1,600 & 23.2\% & 6,910\,$\rightarrow$\,1,600 & 23.2\% \\
SleepEDF & 330,638\,$\rightarrow$\,22,790 & 6.9\% & 331,030\,$\rightarrow$\,22,863 & 6.9\% & 330,374\,$\rightarrow$\,22,861 & 6.9\% \\
HMC & 110,367\,$\rightarrow$\,23,691 & 21.5\% & 110,139\,$\rightarrow$\,23,488 & 21.3\% & 109,665\,$\rightarrow$\,23,715 & 21.6\% \\
TUAB & 324,769\,$\rightarrow$\,76,000 & 23.4\% & 327,890\,$\rightarrow$\,76,280 & 23.3\% & 328,390\,$\rightarrow$\,76,320 & 23.2\% \\
TUEV & 87,239\,$\rightarrow$\,16,409 & 18.8\% & 94,090\,$\rightarrow$\,16,946 & 18.0\% & 93,048\,$\rightarrow$\,16,691 & 17.9\% \\
TUSL & 201\,$\rightarrow$\,201 & 100.0\% & 200\,$\rightarrow$\,200 & 100.0\% & 218\,$\rightarrow$\,218 & 100.0\% \\
EEGDenoiseNet & 10,809\,$\rightarrow$\,80 & 0.7\% & 10,809\,$\rightarrow$\,80 & 0.7\% & 10,809\,$\rightarrow$\,80 & 0.7\% \\
\midrule
\textbf{Total} & 1,397,300\,$\rightarrow$\,191,164 & 13.7\% & 1,406,812\,$\rightarrow$\,191,859 & 13.6\% & 1,405,361\,$\rightarrow$\,191,879 & 13.7\% \\
\end{longtable}
\endgroup
\tabnote{The cap retains at most 40 windows per registered subject--class unit from each training split; validation and test use all available windows. Retention is the after/before ratio. Totals count task entries within each seed, including separate LEMON tasks; seeds reuse overlapping windows.}
\endgroup

\section{Full-FT Duration--Capacity Response}
\label{app:ft-duration-fit}

We fit the shared response to the 12 Full-FT observations in Figure~\ref{fig:data-fractions}: four fractions each for Molecule-B, Atom-B, and Atom-L. Each score averages the same 23 tasks and three downstream seeds at the end of a 50-epoch run. Subset hours are estimated as the retained fraction times the collection duration.

In Equation~\ref{eq:ft-scaling-response}, the shared response is $g(u)=2u/(1+u^2)=[(u+u^{-1})/2]^{-1}$. It approaches $2u$ for small $u$ and $2/u$ for large $u$, with a maximum at $u=1$. We use this bounded, log-symmetric shape to describe the measured duration range. Both tails approach the series-specific level $A_i$.

There are six fitted parameters: three levels $A_i$, one common amplitude $B$, and two peak durations. Weighted least squares profiles out the levels and amplitude while searching for the peaks. We set the peak-search bounds to 1,000--1,000,000 hours for Base and 24,814--49,628 hours for Large. The size exponent is then $\eta=\log(H_L/H_B)/\log(46.3/5.8)$.

\begin{reportfigure}
\centering
\includegraphics[width=\linewidth]{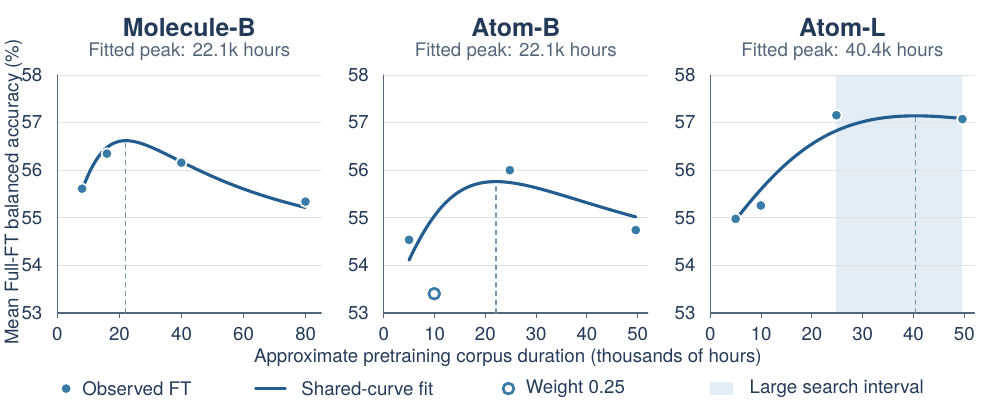}
\caption{Shared Full-FT response with Atom-B's 20\% point assigned weight 0.25. All 12 observations are shown; the hollow point receives reduced weight. Curves share the response shape and amplitude and are drawn only over observed duration ranges. Dashed lines mark fitted peaks; shading marks the imposed Large peak-search interval.}
\label{fig:ft-duration-interpolation}
\end{reportfigure}

\begin{reporttable}
\centering\small
\caption{Sensitivity to the weight of Atom-B's 20\% point. All other points have unit weight. Peaks are in thousands of hours; leave-one-point-out (LOO) RMSE uses all 12 held-out scores, in percentage points.}
\label{tab:ft-duration-sensitivity}
\begin{tabular}{rrrrrr}
\toprule
Weight & Base peak & Large peak & $\eta$ & LOO RMSE & Constant LOO \\
\midrule
1 & 25.22 & 40.81 & 0.232 & 0.919 & 1.094 \\
0.25 & 22.07 & 40.45 & 0.292 & 0.707 & 1.076 \\
0 & 20.45 & 40.09 & 0.324 & 0.680 & 1.080 \\
\bottomrule
\end{tabular}
\end{reporttable}

The displayed fit gives $K=22{,}065.77$ hours, $\eta=0.291689$, and $B=2.8808$ percentage points. The fitted levels $A_i$ are 52.8795, 54.2612, and 53.7392 for Atom-B, Atom-L, and Molecule-B. We use weight 0.25 for this fit and report weights 1 and 0 for sensitivity analysis. The three seeds measure downstream variability for each pretrained checkpoint.

\subsection{Response-shape sensitivity and the inverse relation}
\label{app:duration-reference}

The peak estimates depend on how the FT response is modeled. For comparison, we fit $S_i=a_i-b_i(u-1)^2$ and $S_i=a_i-b_i[\log u]^2$ with separate levels and amplitudes for each series and the same peak-search bounds (Table~\ref{tab:independent-ft-peaks}). These alternatives have eight parameters for 12 observations. The raw-duration quadratic assumes symmetric declines per hour; the log-duration quadratic assumes symmetry per multiplicative change in duration.

\begin{reporttable}
\centering\small
\caption{Alternative response shapes with series-specific amplitudes. Atom-B's 20\% point receives the indicated weight. Peak durations are in thousands of hours.}
\label{tab:independent-ft-peaks}
\begin{tabular}{lrrrr}
\toprule
Response shape & Weight & Base peak & Large peak & $\eta$ \\
\midrule
Quadratic in duration & 1 & 31.64 & 37.62 & 0.083 \\
 & 0.25 & 29.40 & 37.62 & 0.119 \\
 & 0 & 28.27 & 37.62 & 0.138 \\
\addlinespace[2pt]
Quadratic in log duration & 1 & 24.29 & 49.63$^\dagger$ & 0.344 \\
 & 0.25 & 21.92 & 49.63$^\dagger$ & 0.393 \\
 & 0 & 16.80 & 49.63$^\dagger$ & 0.521 \\
\bottomrule
\end{tabular}
\par\smallskip
{\footnotesize $^\dagger$The fitted Large peak reaches the upper search bound.}
\end{reporttable}

An independent raw-duration quadratic places Molecule-B's peak at 37,473 hours, compared with 15,982 hours for its best measured score. Response shape thus has a substantial effect on the inferred peak. The bounded response in Equation~\ref{eq:ft-scaling-response} gives an interior Large peak for all three point weights.

Sharing the amplitude lowers LOO error within the bounded-response family. At weight 0.25, allowing a separate amplitude for each of the three bounded curves gives a LOO RMSE of 1.259 points, compared with 0.707 for the common-amplitude model. Sharing only between Atom-B and Atom-L gives 0.879.

For each LOO prediction, we remove one observation and refit all parameters. The constant comparison fits one level per series with the same training weights and gives an RMSE of 1.076 points. We use LOO for internal model selection, with the same observations guiding the response shape and amplitude-sharing choices. Endpoint omissions require extrapolation; keeping only held-out 20\% and 50\% points gives 0.880 points for the shared response and 1.217 for the constant comparison.

The inverse of the fitted peak relation is
\begin{equation}
P_{\mathrm{ref}}(H)=5.8\times10^6
\left(\frac{H}{K}\right)^{1/\eta},\qquad
\frac{d\log P_{\mathrm{ref}}}{d\log H}=\frac{1}{\eta}.
\label{eq:duration-reference-inverse}
\end{equation}
The fitted inverse exponent is 3.4283; Equation~\ref{eq:data-saturation} uses $1/0.292$ after rounding. Thus a relative change in duration produces a larger relative change in the corresponding parameter estimate. Here $P_{\mathrm{ref}}(H)$ identifies the encoder size whose fitted peak occurs at duration $H$. Selecting a size for maximum FT performance at fixed duration additionally requires estimating how the series levels $A_i$ vary with capacity and collection.

The exponent interpolates the two fitted peak locations for Base and Large. The resulting empirical scaling hypothesis describes FT under the current data collections and 50-epoch schedules, which couple corpus duration to optimization updates. Additional model sizes, intermediate fractions, and matched training exposure would test how this relation generalizes. LP follows a different fraction-dependent response and is analyzed separately in Section~\ref{sec:data-saturation}.

\end{document}